\documentclass[10pt,twocolumn,letterpaper]{article}

\usepackage{iccv}
\usepackage{times}
\usepackage{epsfig}
\usepackage{graphicx}
\usepackage{amsmath}
\usepackage{amssymb}

\usepackage{booktabs}
\usepackage{subcaption}
\usepackage{multirow}
\usepackage{pifont}
\newcommand{\cmark}{\ding{51}}%
\newcommand{\xmark}{\ding{55}}%

\usepackage[pagebackref=true,breaklinks=true,letterpaper=true,colorlinks,bookmarks=false]{hyperref}

\iccvfinalcopy 

\def\iccvPaperID{5504} 

\ificcvfinal\fi

\begin{document}

\title{Robust Object Modeling for Visual Tracking}

\author{Yidong Cai \qquad Jie Liu\thanks{Corresponding author.} \qquad Jie Tang \qquad Gangshan Wu\\
State Key Laboratory for Novel Software Technology, Nanjing University, China\\
{\tt\small yidong\_cai@smail.nju.edu.cn; \{liujie, tangjie, gswu\}@nju.edu.cn}
}

\maketitle
\ificcvfinal\thispagestyle{empty}\fi

\begin{abstract}
   Object modeling has become a core part of recent tracking frameworks. Current popular tackers use Transformer attention to extract the template feature separately or interactively with the search region. However, separate template learning lacks communication between the template and search regions, which brings difficulty in extracting discriminative target-oriented features. On the other hand, interactive template learning produces hybrid template features, which may introduce potential distractors to the template via the cluttered search regions. To enjoy the merits of both methods, we propose a robust object modeling framework for visual tracking (ROMTrack), which simultaneously models the inherent template and the hybrid template features. As a result, harmful distractors can be suppressed by combining the inherent features of target objects with search regions' guidance. Target-related features can also be extracted using the hybrid template, thus resulting in a more robust object modeling framework. To further enhance robustness, we present novel variation tokens to depict the ever-changing appearance of target objects. Variation tokens are adaptable to object deformation and appearance variations, which can boost overall performance with negligible computation. Experiments show that our ROMTrack sets a new state-of-the-art on multiple benchmarks. Code is available at \href{https://github.com/dawnyc/ROMTrack}{https://github.com/dawnyc/ROMTrack}.
\end{abstract}

\section{Introduction}
\label{sec:intro}
Visual object tracking (VOT)~\cite{Staple, BhatJDKF18, FCOT, ATOM, GaloogahiFL17, KCF, LukezicVZMK17, MDNet, Song0WGBZSL018, WangZLWTB21, UpdateNet, DaSiamRPN} is a fundamental task in computer vision, which aims at localizing an arbitrary target in video sequences given its initial status. The occlusion, scale variation, object deformation, and co-occurrence of distractor objects pose a challenge to acquiring an effective tracker in real-world scenarios. Current dominating trackers typically address these problems with a Transformer-based~\cite{AttentionIsAllYouNeed} architecture.

\begin{figure}
    \centering
    \includegraphics[width=\linewidth]{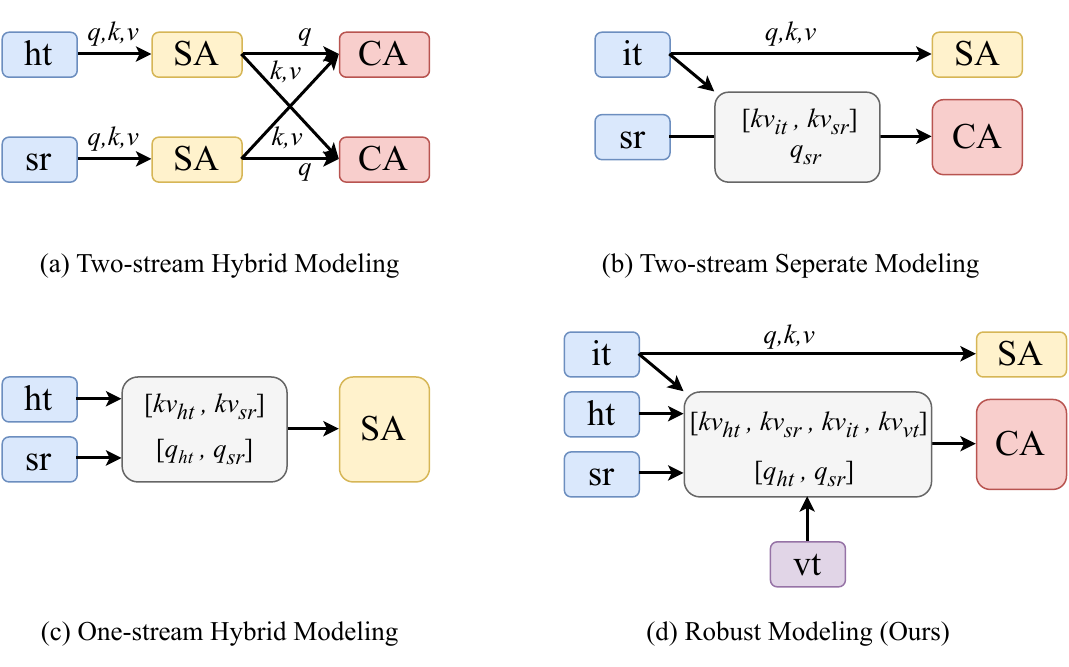}
    \caption{Three typical object modeling methods for template-search feature learning, together with our Robust Modeling design. $ht$, $it$, and $vt$ represent 
    hybrid template, inherent template, and variation tokens, respectively. $sr$ represents the search region. SA and CA denote self-attention and cross-attention, respectively.}
    \label{fig:teaser}
\end{figure}

The core components in a typical Transformer tracking framework are the object modeling blocks. As demonstrated in Figure~\ref{fig:teaser}(a), the two-stream hybrid modeling methods~\cite{TransT,SBT} learn the template feature interactively with the search region via two cross-attention (CA) operations. 
Instead of cross-attention, the one-stream hybrid modeling methods~\cite{SimTrack,OSTrack} in Figure~\ref{fig:teaser}(c) jointly learn the hybrid template feature and search region feature with one self-attention (SA) operation. Different from hybrid template modeling, the two-stream separate modeling~\cite{MixFormer,AiATrack} in Figure~\ref{fig:teaser}(b) keeps an inherent template stream to ensure the purity of template features.
Separate template learning can keep the inherent features of target templates, which prevents interference from search regions. Though suffering from potential distractors, hybrid template learning conducts extensive feature matching between the template and search region, thus allowing mutual guidance for target-oriented feature extraction.

In order to enjoy the merits of separate and hybrid template modeling simultaneously, we propose a robust object modeling framework for visual tracking (named ROMTrack). As shown in Figure~\ref{fig:teaser}(d), our robust modeling scheme involves two kinds of templates, the inherent template $it$ and the hybrid template $ht$. Meanwhile, our scheme also designs the novel variation tokens $vt$. The inherent template applies self-attention separately to enhance its learned feature. Besides, it accepts queries from the hybrid template and the search region features to provide inherent information for discriminative target-oriented feature learning. 
The bottom part of Figure~\ref{fig:teaser}(d) is a hybrid attention that adopts a standard cross-attention operation to enhance the template and search region features with mutual guidance. 
Furthermore, it is well-recognized that tracking is a task suffering from object deformation and appearance variations~\cite{MixFormer,UpdateNet}. We tackle this problem by introducing novel variation tokens to improve robustness. It is observed that the target's motion during a short period is usually smooth but may be accompanied by large changes in appearance~\cite{TransT,Ocean}. The tracker can easily handle smooth motion, but appearance changes are hard to distinguish. Therefore, we generate variation tokens from hybrid template features to leverage appearance information during tracking. Despite the simplicity, our variation tokens perform well with negligible computation.

The main contributions of this work are three-fold: (1) We propose a robust object modeling framework for visual tracking (ROMTrack). It can keep the inherent information of the target template and enables mutual feature matching between the target and the search region simultaneously. (2) We present a neat and effective variation-token design that embeds appearance context during tracking into the attention calculation of hybrid target-search features. (3) The proposed ROMTrack sets a new state-of-the-art performance on six challenging benchmarks, including GOT-10k~\cite{GOT-10k}, LaSOT~\cite{LaSOT}, TrackingNet~\cite{TrackingNet}, LaSOT$_{\textbf{ext}}$~\cite{LaSOT_ext}, OTB100~\cite{OTB100}, and NFS30~\cite{NFS30}.


\section{Related Work}
In this section, we briefly review different visual object tracking methods and the Transformer attention mechanism in general vision tasks.

\noindent\textbf{Visual Object Tracking.}
Early Siamese-based trackers~\cite{SiamFC, SiamBAN, CGACD, CRPN, SiamRPN++, SiameseRPN, RASNet, SiamFC++, SiamAttn, Ocean} first extract the template and search region features separately by a CNN (Convolutional Neural Network) backbone with shared structure and parameters. Then, a correlation-based network is responsible for computing the similarity between the template and the search region. Correlation modeling plays a critical role in tracking networks. However, conventional correlation-based networks do not fully use the global context. Therefore, recent dominating trackers~\cite{TransT, TREG, MixFormer, STMTrack, AiATrack, SwinTrack, TrDiMP, STARK, OSTrack, DTT} introduce stacked Transformer layers for better relation modeling.

The pioneering Transformer tracking method TransT~\cite{TransT} adopts a similar pipeline as Siamese-based trackers, where the lightweight relation modeling network is replaced with relatively heavy Transformer layers. The two-stream attention in TransT enables bi-directional information interaction. 
Unlike TransT, MixFormer~\cite{MixFormer} utilizes the flexibility of attention operations for simultaneous feature extraction and relation modeling. MixFormer also adopts a two-stream attention pipeline but prunes the cross-attention from the target's query to the search area, eliminating potential negative influence from distractors. 
AiATrack~\cite{AiATrack} employs a similar asymmetric scheme, where the search region conducts queries on the target feature while the target only enhances its feature with self-attention blocks. 
In order to bridge a free information flow between the template and search region, OSTrack~\cite{OSTrack} adopts a one-stream attention scheme. It concatenates the flattened template and search region and feeds them into stacked self-attention layers for joint feature learning and relation modeling. However, the extensive feature fusion of self-attention layers may bring interfering information to the target feature due to potential distractors. 
Instead, we propose a robust object modeling scheme that contains an inherent template stream, a variation-token stream, and a bi-directional template-search stream, leading to a more accurate transformer tracker.

\noindent\textbf{Transformer Attention.}
The attention mechanism~\cite{AttentionIsAllYouNeed} has played an increasingly important role in computer vision in the past few years. And recently, in most vision tasks, attention architectures represented by Transformer have obtained impressive performances. To be more specific, Transformer attention is usually helpful for modeling spatial features and temporal relations. For example, the Vision Transformer (ViT)~\cite{ViT} and other following works, including PVT~\cite{PVT}, CVT~\cite{CvT}, and Swin-Transformer~\cite{SwinTransformer} have shown their capacity to aggregate spatial information and benefit many downstream tasks. Transformer attention has also been utilized in visual tracking but has yet to be fully exploited. Most of them focus on designing complex structures. Instead, in this work, we try to explain the potential defects of previous trackers and seek an approach for robust object modeling. We follow the pure Transformer architecture to further explore attention mechanism for visual tracking. 

\begin{figure*}[t]
    \centering
    \includegraphics[width=\linewidth]{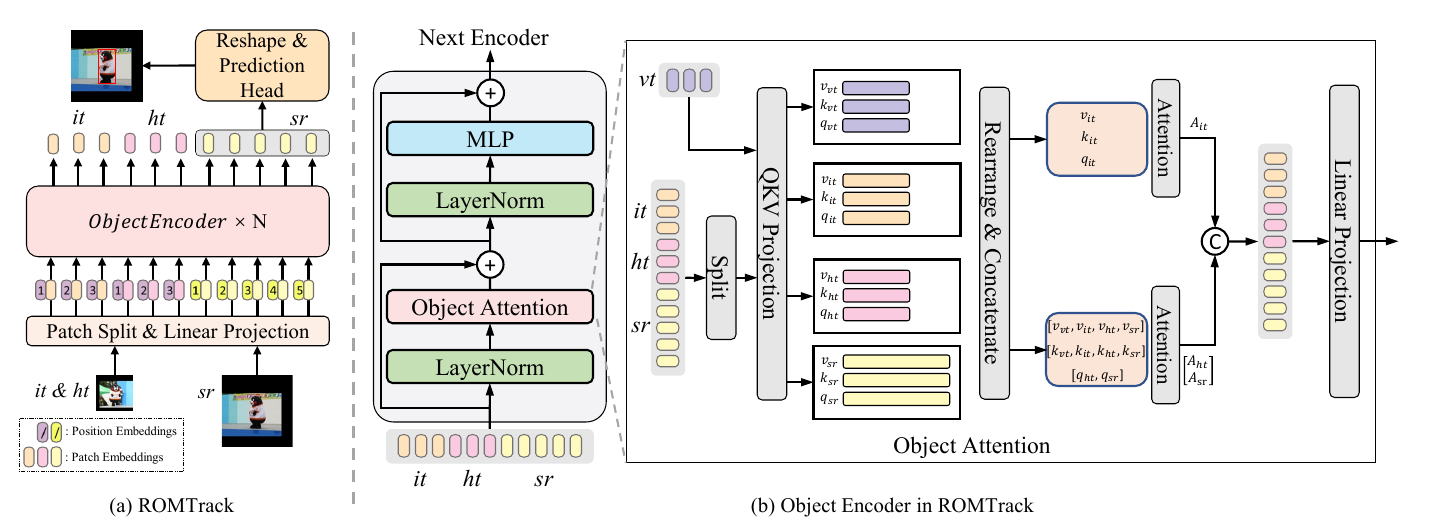}
    \caption{(a) Overview of the proposed ROMTrack framework. The template and search region images are split into patches, and then linearly projected, concatenated, and fed into stacked encoder layers for robust object modeling. $it$, $ht$, and $sr$ denote the inherent template, the hybrid template, and the search region, respectively.
    (b) Architecture of the object encoder layer. $vt$ denotes variation tokens.} 
    \label{fig:overview}
\end{figure*}

\section{Method}
We propose ROMTrack, a robust object modeling network for tracking, in Figure~\ref{fig:overview}. We first give an overview of the proposed ROMTrack architecture and then elaborate on the proposed object encoder. Finally, we give a discussion on our modeling method.

\subsection{Overall Architecture}\label{sec:overall_architecture}
\noindent\textbf{Backbone.} As shown in Figure~\ref{fig:overview}(a), we adopt the vanilla ViT~\cite{ViT} as the backbone. More concretely, we replace the conventional ViT encoder with the proposed object encoder and add a prediction head on the output tokens of the last encoder. The input of ROMTrack is a triplet of images containing a template image pair $(it_{img},ht_{img})\in \mathbb{R}^{3\times H_{t} \times W_{t}}$ and one search region image $sr_{img}\in \mathbb{R}^{3\times H_{sr} \times W_{sr}}$. 
The $it_{img}$ is responsible for learning inherent template features and $ht_{img}$ is accountable for learning hybrid template features.
Following ViT~\cite{ViT}, we split all input images and flatten them into sequences of patches: $it_{p}\in \mathbb{R}^{N_{t} \times 3\cdot P^2}$,
$ht_{p}\in \mathbb{R}^{N_{t} \times 3\cdot P^2}$, and $sr_{p}\in \mathbb{R}^{N_{sr} \times 3\cdot P^2}$, where $P\times P$ is the resolution of each patch, $N_t=H_tW_t/P^2$ and $N_{sr}=H_{sr}W_{sr}/P^2$ are the number of patches of the templates and the search region, respectively. 
Then we generate $D$-dimensional patch embeddings with a linear projection layer. After adding position embeddings, the resulting token sequences are ready for $N$ stacked object encoders. The encoder layer employs robust object modeling to learn discriminative feature representations, which will be elaborated in Section~\ref{sec:object_encoder}.

\noindent\textbf{Prediction Head.} As pointed out in previous work~\cite{MixFormer}, corner-based~\cite{CornerNet} localization heads may have a bad effect on the modeling capacity of deeper transformer encoders. 
Consequently, We adopt a fully convolutional center-based~\cite{CenterNet} localization head to estimate the bounding box of tracked objects, which consists of $L$ stacked Conv-BN-ReLU layers. 
Specifically, the target classification score map $C\in[0,1]^{\frac{H_{sr}}{P}\times\frac{W_{sr}}{P}}$, the local offset map $O\in[0,1]^{2\times\frac{H_{sr}}{P}\times\frac{W_{sr}}{P}}$, and the normalized bounding box size map $S\in[0,1]^{2\times\frac{H_{sr}}{P}\times\frac{W_{sr}}{P}}$ are generated by the center head. 
Finally, the position with the highest classification score in $C$ is considered the target position and the target bounding box can be calculated using $O$ and $S$.

The classification branch is supervised using Gaussian weighted focal loss~\cite{CornerNet} during training. 
Specifically, given a ground truth target center $\hat{c}$ and the corresponding position $\tilde{c} = [\tilde{c}_x, \tilde{c}_y]$ in feature map, the ground truth heatmap can be formulated as $\hat{C}_{xy} = e^{-\frac{(x-\tilde{c}_x)^2 + (y-\tilde{c}_y)^2}{2\sigma^2}}$, where $\sigma$ is a standard deviation adaptive to object size. 
So the Gaussian weighted focal loss is employed as follows:
\begin{equation}
  \begin{aligned}
    L_{cls} = -\sum_{xy}& [{\mathbb{I}(\hat{C}_{xy} = 1)}(1-{C}_{xy})^{\alpha}\log({C}_{xy}) \\
    & + (1-\hat{{C}}_{xy})^{\beta}({C}_{xy})^{\alpha}\log(1-{C}_{xy})],
  \end{aligned}
  \label{eq:focalloss}
\end{equation}
where {$\mathbb{I}(\cdot)$ is the indicator function,} $\alpha$ and $\beta$ are hyper-parameters, and we set them to 2 and 4 following~\cite{CornerNet,OSTrack}.

As for the bounding box regression branch, $L_1$ loss and $GIoU$ loss are adopted. 
Generally, We set different weights for different losses: $\lambda_{L_1} = 5$, $\lambda_{giou} = 2$ and $\lambda_{cls} = 1$.
And both training stages share the same loss function as follows:
\begin{equation}
  \begin{aligned}
    L_{total} = \lambda_{L_1}L_1 + \lambda_{giou}L_{giou} + \lambda_{cls}L_{cls}.
  \end{aligned}
  \label{eq:totalloss}
\end{equation}

\subsection{Object Encoder}\label{sec:object_encoder}
The proposed object encoder in Figure~\ref{fig:overview}(b) contains two critical components, \ie, variation tokens and robust object modeling. 
Before describing the principles of robust object modeling, we first explain our design of variation tokens.

\begin{figure}[t]
    \centering
    \includegraphics[width=\linewidth]{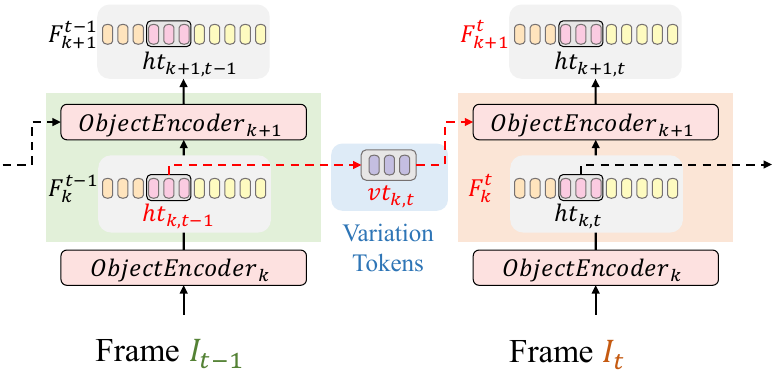}
    \caption{Schema of the proposed variation-token design. }
    \label{fig:variation_token}
\end{figure}

\noindent\textbf{Variation Tokens.} 
Variation tokens are the embedding of contextual appearance changes of the target object, which helps to tackle the problem of object deformation and appearance variations. As shown in Figure~\ref{fig:variation_token}, the variation tokens $vt$ are generated after each object encoder and encode the variation of appearance context from the search region, which will be further demonstrated later. 
The generation and usage of variation tokens can be formulated as follows:
\begin{align}
    vt_{k,t} &= ht_{k,t-1}\label{eq:1},\\
    F_{k+1}^{t} &= ObjectEncoder_{k+1}(\text{Concat}(vt_{k,t},F_{k}^{t}))\label{eq:2},
\end{align}
where $F$ represents the output features, $k$ is the encoder index and $t$ denotes the $t$-th frame. 
So $F_{k}^{t}$ is the output features of $k$-th encoder in frame $I_{t}$, and $ht_{k,t}$ is the hybrid template part of $F_{k}^{t}$, which incorporates appearance information from the search region.

Equation~\ref{eq:1} indicates that we reserve hybrid template tokens of frame $I_{t-1}$ as the input to frame $I_{t}$, because appearance variations of the target object in $I_{t}$ relative to $I_{1}$ are encoded at the feature level of these tokens. Furthermore, Equation~\ref{eq:2} aims to embed the variation of object appearance into the network. Specifically, it feeds the variation tokens $vt_{k,t}$ together with the output features $F_k^t$ to the $(k+1)$-th encoder when tracking the $t$-th frame. The MACs are negligible because the construction of variation tokens only includes embedding assignments (Equation~\ref{eq:1}). Meanwhile, the employment of variation tokens is just a combination of token concatenation and a series of lightweight cross-attention operations related to $vt_{k,t}$ (Equation~\ref{eq:2}).

\noindent\textbf{Robust Object Modeling.}
One-stream hybrid modeling enables extensive bi-directional information flows between the template-search image pairs, and discriminative target-oriented features can be dynamically extracted by mutual guidance. 
However, excessive communications may suffer from tracking failures and background clutters. 
Two-stream separate modeling can keep a separate template stream to avoid negative influences from potential distractors, but its extracted template features are inadequate to object deformations and appearance changes. 

To address these problems, we propose a robust object modeling method. As shown in Figure~\ref{fig:overview}(b), the input of object attention consists of four parts, \ie, the inherent template $it \in\mathbb{R}^{N_{t}\times D}$, the hybrid template $ht \in\mathbb{R}^{N_{t}\times D}$, the search region $sr \in\mathbb{R}^{N_{sr}\times D}$, and the variation tokens $vt\in\mathbb{R}^{N_{t}\times D}$.
Following the conventional attention block, we use a linear projection layer to produce the $d$-dimentional $(query, key, value)$ triplet. For example, the triplet for $it$ is $(q_{it}, k_{it}, v_{it})$. Then we conduct self-attention on $it$ to learn pure template features:
\begin{equation}
    A_{it} = \text{Softmax}(\frac{q_{it}k_{it}^{T}}{\sqrt{d}})v_{it},
\end{equation}
where $A_{it}$ represents the output of the self-attention operation. The inherent template feature is enhanced through self-attention, eliminating interference from the search region. Meanwhile, the hybrid template feature and the search region feature are learned via a cross-attention operation.
Let $(q_{z},k_{z},v_{z})$ denote the $(query,key,value)$ triplet of the cross-attention, where $q_{z}$, $k_{z}$, and $v_{z}$ are defined as follows:
\begin{align}
    q_{z} &= [q_{ht}, q_{sr}],\label{eq:qm}\\
    k_{z} &= [k_{vt}, k_{it}, k_{ht}, k_{sr}],\label{eq:km}\\
    v_{z} &= [v_{vt}, v_{it}, k_{ht}, v_{sr}]\label{eq:vm}.
\end{align}

Equation~\ref{eq:qm} - \ref{eq:vm} show that the inputs of cross-attention are a rearrangement and concatenation (denoted by $[\dots]$) of the search region features (indicated by subscript $sr$), the two types of template features (indicated by subscript $ht$ and $it$), and the variation token features (indicated by subscript $vt$). The output of cross-attention operation $A_{z}$ can be obtained via:
\begin{equation}
    A_{z} = \text{Softmax}(\frac{q_{z}k_{z}^{T}}{\sqrt{d}})v_{z}.
\end{equation}

The hybrid template feature and search region feature inside $A_z$ get enhanced by fusing informative features from the inherent template and variation tokens. As a result, the network is able to obtain the information of both the original target (\ie, object in the first frame) and the ever-changing target (\ie, object in the $(t-1)$-th frame) when tracking the $t$-th frame.

For further explanation, we conduct a more in-depth analysis below.
Let $M_{z}$ be the correlation map calculated in the cross-attention, then $M_{z}$ can be written as: 
\begin{equation} 
\begin{split}
    M_z =& \text{Softmax}(\frac{q_{z}k_{z}^{T}}{\sqrt{d}})\\
        =& \text{Softmax}(\frac{\begin{bmatrix} q_{ht}k_{vt}^T & q_{ht}k_{it}^T & q_{ht}k_{ht}^T & q_{ht}k_{sr}^T \\ q_{sr}k_{vt}^T & q_{sr}k_{it}^T & q_{sr}k_{ht}^T & q_{sr}k_{sr}^T \end{bmatrix}}{\sqrt{d}})\\
        \triangleq & \begin{bmatrix} M_{ht,vt} & M_{ht,it} & M_{ht,ht} & M_{ht,sr} \\ M_{sr,vt} & M_{sr,it} & M_{sr,ht} & M_{sr,sr} \end{bmatrix},
\end{split}\raisetag{2\baselineskip}
\label{eq:M}
\end{equation}
where $M_{a,b}$ is a measure of similarity between $a$ and $b$, \eg, $M_{ht,sr}$ refers to the similarity between the hybrid template and search region. 
Based on Equation~\ref{eq:M}, the attention output $A_z$ can be rewritten as:
\begin{equation}
\label{eq:A}
  \begin{split}
    A_{z} =& M_{z}v_z = \text{Softmax}(\frac{q_z k_z^T}{\sqrt{d}})v_z\\
    =& \begin{bmatrix} M_{ht,vt} & M_{ht,it} & M_{ht,ht} & M_{ht,sr} \\ M_{sr,vt} & M_{sr,it} & M_{sr,ht} & M_{sr,sr} \end{bmatrix} \begin{bmatrix} v_{vt} \\ v_{it} \\ v_{ht} \\ v_{sr} \end{bmatrix}\\
    =& \begin{bmatrix} M_{ht,vt}v_{vt} + M_{ht,it}v_{it} + M_{ht,ht}v_{ht} + M_{ht,sr}v_{sr} \\ M_{sr,vt}v_{vt} + M_{sr,it}v_{it} + M_{sr,ht}v_{ht} + M_{sr,sr}v_{sr} \end{bmatrix}\\
    \triangleq & \begin{bmatrix} A_{ht}\\ A_{sr} \end{bmatrix},
  \end{split}\raisetag{5\baselineskip}
\end{equation}
where $A_{ht}$ and $A_{sr}$ denote the generated hybrid template and search region features, respectively. 
It is easy to figure out that both $A_{ht}$ and $A_{sr}$ aggregate information from the inherent template (\eg, $M_{ht,it}v_{it}$ and $M_{sr,it}v_{it}$) and variation tokens (\eg, $M_{ht,vt}v_{vt}$ and $M_{sr,vt}v_{vt}$) to enhance their features. 

Recall that in Figure~\ref{fig:variation_token}, we use the hybrid template $ht_{k,t-1}$ to generate the variation tokens $vt_{k,t}$ to provide variation information of contextual appearance for the next frame.
This is reasonable because the $M_{ht,sr}v_{sr}$ term in $A_{ht}$ has incorporated the feature of search region into the hybrid template, helping the output hybrid template tokens to capture current information of the search region. In other words, the appearance information of the target object in both the first frame and the current frame is incorporated into the hybrid template tokens, making them sensitive to contextual appearance changes. 

Therefore, we can cache the hybrid template as variation tokens in the next frame to leverage appearance information during tracking. Overall, with the variation-token design, the feature extraction and information integration process are unified in the proposed object modeling framework.

\subsection{Discussions}\label{sec:discussions}
\noindent\textbf{Necessity of Hybrid Template.}
The hybrid template serves two primary purposes. The first is to conduct extensive feature matching between the template and search region, thus allowing mutual guidance for target-oriented feature extraction. The second is to encode the variation of appearance context by interacting with the search region, which helps variation tokens model appearance changes of objects between adjacent frames. Further analysis is conducted in Section~\ref{sec:Ablation}.

\noindent\textbf{Training and Inference.}
The training process contains two stages. 
In the first stage, we follow the standard training recipe of mainstream trackers~\cite{TransT,MixFormer,STARK} to train our ROMrack without variation tokens, \ie, only with the inherent and hybrid templates.
In the second stage, we add variation tokens into training by sampling two search regions in consecutive frames of the same sequence to model the appearance variations between them. 
For inference, only the initial template and the cropped search region are fed into the ROMTrack pipeline to produce the target bounding box. 
The initial template serves as the input for both inherent and hybrid templates. 
During the tracking procedure, the variation tokens are obtained per frame and employed for subsequent tracking.

\begin{table*}
  \centering
  \resizebox{\linewidth}{!}{
  \begin{tabular}{c|c|ccc|ccc|ccc|ccc}
    \toprule
    \multirow{2}{*}{Method} & \multirow{2}{*}{Source} & \multicolumn{3}{c|}{GOT-10k*} & \multicolumn{3}{c|}{LaSOT} & \multicolumn{3}{c|}{TrackingNet} & \multicolumn{3}{c}{LaSOT$_{\textbf{ext}}$} \\
    \cline{3-14}
     & & $AO(\%)$ & $SR_{0.5}(\%)$ & $SR_{0.75}(\%)$ & $AUC(\%)$ & $P_{Norm}(\%)$ & $P(\%)$ & $AUC(\%)$ & $P_{Norm}(\%)$ & $P(\%)$ & $AUC(\%)$ & $P_{Norm}(\%)$ & $P(\%)$ \\
    \midrule
    \textbf{ROMTrack} & Ours & \color{red}\textbf{72.9} & \color{red}\textbf{82.9} & \color{red}\textbf{70.2} & \color{red}\textbf{69.3} & 78.8 & \color{red}\textbf{75.6} & \color{red}\textbf{83.6} & \color{red}\textbf{88.4} & \color{red}\textbf{82.7} & \color{red}\textbf{48.9} & \color{red}\textbf{59.3} & \color{red}\textbf{55.0} \\
    \hline
    SwinTrack-T-224~\cite{SwinTrack} & NIPS22 & \color{blue}\textbf{71.3} & \color{blue}\textbf{81.9} & 64.5 & 67.2 & - & 70.8 & 81.1 & - & 78.4 & 47.6 & - & 53.9 \\
    OSTrack-256~\cite{OSTrack} & ECCV22 & 71.0 & 80.4 & \color{blue}\textbf{68.2} & 69.1 & 78.7 & \color{blue}\textbf{75.2} & 83.1 & 87.8 & 82.0 & 47.4 & 57.3 & 53.3 \\
    OSTrack-256(w/o CE)~\cite{OSTrack} & ECCV22 & 71.0 & 80.3 & 68.2 & 68.7 & 78.1 & 74.6 & 82.9 & 87.5 & 81.6 & - & - & - \\
    AiATrack~\cite{AiATrack} & ECCV22 & 69.6 & 80.0 & 63.2 & 69.0 & \color{red}\textbf{79.4} & 73.8 & 82.7 & 87.8 & 80.4 & 46.8 & 54.4 & \color{blue}\textbf{54.2} \\
    SimTrack-B/16~\cite{SimTrack} & ECCV22 & 68.6 & 78.9 & 62.4 & \color{blue}\textbf{69.3} & 78.5 & 74.0 & 82.3 & 86.5 & - & - & - & - \\
    Unicorn~\cite{Unicorn} & ECCV22 & - & - & - & 68.5 & 76.6 & 74.1 & 83.0 & 86.4 & \color{blue}\textbf{82.2} & - & - & - \\
    MixFormer-22k~\cite{MixFormer} & CVPR22 & 70.7 & 80.0 & 67.8 & 69.2 & 78.7 & 74.7 & \color{blue}\textbf{83.1} & \color{blue}\textbf{88.1} & 81.6 & - & - & - \\
    MixFormer-1k~\cite{MixFormer} & CVPR22 & 71.2 & 79.9 & 65.8 & 67.9 & 77.3 & 73.9 & 82.6 & 87.7 & 81.2 & - & - & - \\
    ToMP50~\cite{ToMP} & CVPR22 & - & - & - & 67.6 & 78.0 & 72.2 & 81.2 & 86.2 & 78.6 & 45.4 & 57.6 & - \\
    ToMP101~\cite{ToMP} & CVPR22 & - & - & - & 68.5 & \color{blue}\textbf{79.2} & 73.5 & 81.5 & 86.4 & 78.9 & 45.9 & \color{blue}\textbf{58.1} & - \\
    SBT-large~\cite{SBT} & CVPR22 & 70.4 & 80.8 & 64.7 & 66.7 & - & 71.1 & - & - & - & - & - & - \\
    KeepTrack~\cite{KeepTrack} & ICCV21 & - & - & - & 67.1 & 77.2 & 70.2 & - & - & - & \color{blue}\textbf{48.2} & 58.0 & - \\
    STARK~\cite{STARK} & ICCV21 & 68.8 & 78.1 & 64.1 & 67.1 & 77.0 & - & 82.0 & 86.9 & - & - & - & - \\
    DTT~\cite{DTT} & ICCV21 & 63.4 & 74.9 & 51.4 & 60.1 & - & - & 79.6 & 85.0 & 78.9 & - & - & - \\
    TransT~\cite{TransT} & CVPR21 & 67.1 & 76.8 & 60.9 & 64.9 & 73.8 & 69.0 & 81.4 & 86.7 & 80.3 & 45.1 & 51.3 & 51.2 \\
    TrDiMP~\cite{TrDiMP} & CVPR21 & 67.1 & 77.7 & 58.3 & 63.9 & - & 61.4 & 78.4 & 83.3 & 73.1 & - & - & - \\
    LTMU~\cite{LTMU} & CVPR20 & - & - & - & 57.2 & - & 57.2 & - & - & - & 41.4 & 49.9 & 47.3 \\
    SiamR-CNN~\cite{SiamR-CNN} & CVPR20 & 64.9 & 72.8 & 59.7 & 64.8 & 72.2 & - & 81.2 & 85.4 & 80.0 & - & - & - \\
    Ocean~\cite{Ocean} & ECCV20 & 61.1 & 72.1 & 47.3 & 56.0 & 65.1 & 56.6 & - & - & - & - & - & - \\
    DiMP~\cite{DiMP} & ICCV19 & 61.1 & 71.7 & 49.2 & 56.9 & 65.0 & 56.7 & 74.0 & 80.1 & 68.7 & 39.2 & 47.6 & 45.1 \\
    SiamRPN++~\cite{SiamRPN++} & CVPR19 & 51.7 & 61.6 & 32.5 & 49.6 & 56.9 & 49.1 & 73.3 & 80.0 & 69.4 & 34.0 & 41.6 & 39.6 \\
    MDNet~\cite{MDNet} & CVPR16 & 29.9 & 30.3 & 9.9 & 39.7 & 46.0 & 37.3 & 60.6 & 70.5 & 56.5 & 27.9 & 34.9 & 31.8 \\
    SiamFC~\cite{SiamFC} & ECCV16 & 34.8 & 35.3 & 9.8 & 33.6 & 42.0 & 33.9 & 57.1 & 66.3 & 53.3 & 23.0 & 31.1 & 26.9 \\
    \midrule
    \multicolumn{14}{l}{\textit{Trackers with Higher Resolution or Larger Model}} \\
    \midrule
    \textbf{ROMTrack-384} & Ours  & \color{red}\textbf{74.2} & \color{red}\textbf{84.3} & \color{red}\textbf{72.4} & \color{red}\textbf{71.4} & \color{red}\textbf{81.4} & \color{red}\textbf{78.2} & \color{red}\textbf{84.1} & \color{red}\textbf{89.0} & \color{red}\textbf{83.7} & \color{red}\textbf{51.3} & \color{red}\textbf{62.4} & \color{red}\textbf{58.6} \\
    \hline
    SwinTrack-B-384~\cite{SwinTrack} & NIPS22 & 72.4 & 80.5 & 67.8 & \color{blue}\textbf{71.3} & - & 76.5 & \color{blue}\textbf{84.0} & - & 82.8 & 49.1 & - & 55.6 \\
    OSTrack-384~\cite{OSTrack} & ECCV22 & \color{blue}\textbf{73.7} & \color{blue}\textbf{83.2} & \color{blue}\textbf{70.8} & 71.1 & \color{blue}\textbf{81.1} & \color{blue}\textbf{77.6} & 83.9 & 88.5 & \color{blue}\textbf{83.2} & \color{blue}\textbf{50.5} & \color{blue}\textbf{61.3} & \color{blue}\textbf{57.6} \\
    SimTrack-L/14~\cite{SimTrack} & ECCV22 & 69.8 & 78.8 & 66.0 & 70.5 & 79.7 & 76.2 & 83.4 & 87.4 & - & - & - & - \\
    MixFormer-L~\cite{MixFormer} & CVPR22 & - & - & - & 70.1 & 79.9 & 76.3 & 83.9 & \color{blue}\textbf{88.9} & 83.1 & - & - & - \\
    \bottomrule
  \end{tabular}
  }
  \caption{Comparison with state-of-the-art on four large-scale benchmarks: GOT-10k, LaSOT, TrackingNet, LaSOT$_{\textbf{ext}}$. The best two results are shown in \textcolor{red}{\textbf{red}} and \textcolor{blue}{\textbf{blue}} fonts. * denotes the model trained with only GOT-10k train split.}
  \label{tab:largescaledataset}
\end{table*}

\begin{table*}
  \centering
  \resizebox{\linewidth}{!}{
  \begin{tabular}{c|ccccccccccc|cc}
    \toprule
     & SiamRPN++ & PrDiMP & SuperDiMP & TransT & STARK & KeepTrack & RTS & ToMP & MixFormer-L & OSTrack-384 & AiATrack & ROMTrack & ROMTrack-384 \\
     & ~\cite{SiamRPN++} & ~\cite{PrDiMP} & ~\cite{VOT2018} & ~\cite{TransT} & ~\cite{STARK} & ~\cite{KeepTrack} & ~\cite{RTS} & ~\cite{ToMP} & ~\cite{MixFormer} & ~\cite{OSTrack} & ~\cite{AiATrack} & (Ours) & (Ours) \\
    \midrule
    OTB100 & 69.6 & 69.6 & 70.1 & 69.4 & 68.5 & 70.9 & - & 70.1 & 70.4 & - & 69.6 & \color{red}\textbf{71.4} & \color{blue}\textbf{70.9}\\
    NFS30 & 50.3 & 63.5 & 64.8 & 65.7 & 65.2 & 66.4 & 65.4 & 66.7 & - & 66.5 & 67.9 & \color{blue}\textbf{68.0} & \color{red}\textbf{68.8} \\
    \bottomrule
  \end{tabular}
  }
  \caption{Comparison with state-of-the-art trackers on two small-scale benchmarks: OTB100 and NFS30. Results are compared in terms of AUC($\%$) score. The best two results are shown in \textcolor{red}{\textbf{red}} and \textcolor{blue}{\textbf{blue}} fonts.}
  \label{tab:smallscaledataset}
\end{table*}

\section{Experiments}
\subsection{Implementation Details}
Our trackers are implemented using Python 3.6.13 and PyTorch 1.7.1. The models are trained on 8 Tesla V100 GPUs, and we test the inference speed on a single NVIDIA1080Ti GPU.

\noindent \textbf{Model.} We adopt the vanilla ViT-Base~\cite{ViT} model pretrained with MAE~\cite{MAE} on ImageNet~\cite{ImageNet} as the backbone of our ROMTrack. 
All the input images are split into $16\times 16$ patches. 
As for the prediction head, we adopt a lightweight FCN consisting of 4 stacked Conv-BN-ReLU layers for each output. 
To build an efficient tracker, we adopt a smaller image resolution than other trackers~\cite{MixFormer,AiATrack,STARK}. Namely, the sizes of the template and search images are $128\times 128$ pixels and $256\times 256$ pixels, respectively. 
Furthermore, to verify the scalability of our proposed ROMTrack, we also provide an implementation with a higher resolution called ROMTrack-384, and the sizes of the template and search images are $192\times 192$ pixels and $384\times 384$ pixels. 

\noindent \textbf{Training.} The training splits of COCO~\cite{COCO}, GOT-10k~\cite{GOT-10k}, LaSOT~\cite{LaSOT}, and TrackingNet~\cite{TrackingNet} are used for training. While for the GOT-10k test, we follow the one-shot protocol by only using the GOT-10k train split for training. 
The training process of ROMTrack consists of two stages: the first 300 epochs are for the backbone and head, and the extra 100 are to merge the variation tokens into our architecture.
For data augmentations, horizontal flip and brightness jittering are used following the convention~\cite{MixFormer,STARK,OSTrack}.
We train the ROMTrack using AdamW~\cite{AdamW} with weight decay set to $10^{-4}$. 
For the first stage, the learning rate is initialized as $4\times 10^{-4}$ and decreased to $4\times 10^{-5}$ at the epoch of 240.
For the second stage, the learning rate is initialized as $4\times 10^{-5}$ and decreased to $4\times 10^{-6}$ at the epoch of 80.

\noindent \textbf{Inference.} 
We adopt the Hanning window penalty to utilize positional prior in tracking following the common practice~\cite{TransT,OSTrack,Ocean}. To be more specific, the classification map $C$ is multiplied by the Hanning window with the same size to generate confidence scores, and we simply select the prediction box with the highest confidence score as result. 

\subsection{Comparison with State-of-the-art Trackers}
We compare our ROMTrack with state-of-the-art (SOTA) trackers on six different benchmarks, including four well-known large-scale benchmarks and two commonly used small-scale benchmarks. Results on other datasets are available in Appendix.

\noindent \textbf{GOT-10k.} GOT-10k~\cite{GOT-10k} is a large-scale dataset containing more than 10000 video segments of real-world moving objects. The object classes between train and test sets are zero-overlapped. We follow the one-shot protocol to only train our model on the GOT-10k training split and evaluate the results through the evaluation server. 
As presented in Table~\ref{tab:largescaledataset}, ROMTrack improves all metrics by a large margin, \eg, 1.6$\%$ in AO compared with SwinTrack-T-224 and 2$\%$ in SR$_{0.75}$ compared with OSTrack-256, which indicates the capability in accurate discrimination and localization of objects. 
Furthermore, our higher resolution model ROMTrack-384 sets a new SOTA on the GOT-10k test split, demonstrating that our method has excellent potential to track objects of unseen classes by robust object modeling.

\noindent \textbf{LaSOT.} LaSOT~\cite{LaSOT} is a large-scale, long-term tracking benchmark containing 1400 video sequences: 1120 for training and 280 for testing. We evaluate our ROMTrack on the test set to compare with previous SOTA trackers.
As reported in Table~\ref{tab:largescaledataset}, our ROMTrack shows more accurate and balanced performance, surpassing both OSTrack and MixFormer in all three metrics. Specifically, our higher resolution model ROMTrack-384 establishes a new state-of-the-art on AUC of 71.4$\%$. The result demonstrates that our approach benefits the long-term tracking scenarios.
More analysis of the performance improvements on the LaSOT dataset can be found in Appendix.

\begin{table}
    \centering
    \resizebox{\linewidth}{!}{
    \begin{tabular}{c|ccc|cc}
    \toprule
      Method   &  Speed (FPS) & MACs (G) & Params (M) & LaSOT AUC(\%) & GOT-10k* AO(\%) \\
    \midrule
      OSTrack-256 (w/o CE)~\cite{OSTrack} & 65 & 29.0 & 92.1  & 68.7 & 71.0\\
      MixFormer-22k~\cite{MixFormer} & 25 & 23.0 & 35.6 & 69.2 & 70.7\\
      \textbf{ROMTrack} & 62 & 34.5 & 92.1 & \textbf{69.3} & \textbf{72.9}\\
      \midrule
      OSTrack-384 (w/o CE) & 29 & 65.3 & 92.1  & 71.0 & 73.7\\
      MixFormer-L & 18 & 127.8 & 183.9 & 70.1 & -\\
      \textbf{ROMTrack-384} & 28 & 77.7 & 92.1 & \textbf{71.4} & \textbf{74.2}\\
    \bottomrule
    \end{tabular}
    }
    \caption{Comparison of inference speed, MACs, and Params. We include the results of OSTrack without candidate elimination (w/o CE) here for a fair speed comparison. * denotes the model trained with only GOT-10k train split.}
    \label{tab:complexity}
\end{table}

\noindent \textbf{TrackingNet.} TrackingNet~\cite{TrackingNet} is a large-scale short-term tracking benchmark that provides more than 30000 video sequences with over 14 million boxes. The test split of TrackingNet contains 511 sequences without publicly available ground truth and covers diverse target classes and scenes.
We submit the tracking results to the official evaluation server and make comparisons with previous SOTA trackers in Table~\ref{tab:largescaledataset}. 
The results show that our ROMTrack obtains 83.6$\%$ in AUC and 88.4$\%$ in NP, outperforming previous SOTA MixFormer-22k appreciably. It demonstrates that our approach also benefits short-term visual tracking.

\noindent \textbf{LaSOT$_{\textbf{ext}}$.} LaSOT$_{\textbf{ext}}$~\cite{LaSOT_ext} is an extension of the LaSOT dataset, which contains 150 videos of 15 new object classes. These recently proposed sequences are pretty challenging due to the existence of many similar distractors in the video. 
The results in Table~\ref{tab:largescaledataset} show that our ROMTrack surpasses all other trackers with a large margin and achieves the top-ranked performance on NP of 59.3$\%$, surpassing ToMP by 1.2$\%$. 
Our higher resolution model ROMTrack-384 also outperforms previous trackers by a large margin in all three metrics, setting a new state-of-the-art on LaSOT$_{\textbf{ext}}$, which indicates that our tracker not only has a remarkable generalization ability of unseen classes but also has a robust discrimination ability of similar distractors.

\noindent \textbf{NFS30 and OTB100.} Finally, we report our results on two additional small-scale benchmarks: NFS30~\cite{NFS30} and OTB100~\cite{OTB100}. As shown in Table~\ref{tab:smallscaledataset}, our ROMTrack and ROMTrack-384 have good performances on both benchmarks, establishing SOTA performances. It further indicates the generality of our method.

\noindent\textbf{Speed, MACs and Params.}
We compare the inference speed, MACs, and Params with state-of-the-art trackers in Table~\ref{tab:complexity}. We include the results of OSTrack~\cite{OSTrack} without candidate elimination (w/o CE) for a fair speed comparison. Our ROMTrack can run in real-time at more than 60 FPS, which is on par with OSTrack-256 (w/o CE).
Besides, ROMTrack is 2.5$\times$ faster than MixFormer~\cite{MixFormer}, demonstrating the effectiveness of our robust object modeling. 
For a larger input resolution, our ROMTrack-384 can also achieve comparable speed with OSTrack-384 (w/o CE) and is much faster than MixFormer-L. Moreover, our ROMTrack-384 outperforms MixFormer-L with only 61\% and 50\% of its MACs and Params, respectively.

\noindent\textbf{Discussions.}
Since OSTrack~\cite{OSTrack} also adopts MAE pre-train, we would like to compare with it.
OSTrack employs the one-stream hybrid modeling strategy mentioned in Figure~\ref{fig:teaser}(c), together with a Candidate Elimination (CE) Module, which brings a considerable performance boost (shown in Table~\ref{tab:largescaledataset}). Differently, our ROMTrack does not employ the CE strategy, and it focuses on the robust object modeling mentioned in Figure~\ref{fig:teaser}(d). 
As Table~\ref{tab:largescaledataset} shown, our ROMTrack outperforms  OSTrack-256 (w/o CE) by a large margin even on the challenging LaSOT dataset (+0.6\% AUC), indicating that the proposed robust object modeling is preferable for more accurate discrimination and localization of objects.

\subsection{Ablation Study and Analysis}
\label{sec:Ablation}

We perform a detailed ablation study and an extensive analysis to verify the effectiveness of our method.

\noindent \textbf{Study on Inherent Template and Hybrid Template.}
The most important part of our ROMTrack is the object encoder which unifies the object modeling of template features and search region features. To further verify the effectiveness of the object encoder, we conduct experiments to analyze the performance of the inherent template and the hybrid template designed in our object encoder. 
Specifically, we remove the variation tokens from our object encoder to form a model called ROMTrack (w/o vt), which remains the inherent template and the hybrid template. And then, we compare it with two other commonly used approaches: separate template modeling (STM) and hybrid template modeling (HTM). 
Note that HTM is a reproduction of OSTrack-256 (w/o CE) to make a fair comparison. We also include the results of OSTrack-256 (w/o CE) for more comprehensive comparisons.
The architecture comparison can be found in Table~\ref{tab:ablation}(a).
For fairness, we implement STM, HTM, and ROMTrack (w/o vt) under the same framework and experimental settings. 
As demonstrated in Table~\ref{tab:ablation}(b), our ROMTrack (w/o vt) performs best in all metrics on benchmark datasets, showing the superiority of our robust object modeling.
Moreover, ROMTrack (w/o vt) achieves impressive performances on LaSOT$_{\textbf{ext}}$ dataset, surpassing HTM and STM by 2.1\% and 1.5\% in AUC separately.
In addition, we provide visualizations of attention maps and features for different modeling methods in the Appendix.

\begin{table*}
\begin{subtable}{.33\linewidth}
    \centering
    \resizebox{\linewidth}{!}{
    \begin{tabular}{c|c|c}
    \toprule
    \multirow{2}{*}{Method} & Inherent Template & Hybrid Template \\
     & ($it$) & ($ht$)\\
    \midrule
      OSTrack-256 (w/o CE) & \xmark & \cmark\\
    \midrule
       STM     &   \cmark      & \xmark\\
       HTM     &   \xmark      & \cmark\\
    \midrule
       ROMTrack (w/o vt)  &   \cmark      & \cmark\\
    \bottomrule
    \end{tabular}
    }
    \caption{}
    \end{subtable}
    \hfill
  \begin{subtable}{.665\linewidth}
  \centering
  \resizebox{\linewidth}{!}{
  \begin{tabular}{c|ccc|ccc|ccc}
    \toprule
    \multirow{2}{*}{Method} & \multicolumn{3}{c|}{LaSOT} & \multicolumn{3}{c|}{LaSOT$_{\textbf{ext}}$} & \multicolumn{3}{c}{GOT-10k*} \\
    \cline{2-10}
     &  $AUC(\%)$ & $P_{Norm}(\%)$ & $P(\%)$ & $AUC(\%)$ & $P_{Norm}(\%)$ & $P(\%)$ & $AO(\%)$ & $SR_{0.5}(\%)$ & $SR_{0.75}(\%)$ \\
    \midrule
    OSTrack-256 (w/o CE) & 68.7 & 78.1 & 74.6 & - & - & - & 71.0 & 80.3 & 68.2\\
    \midrule
    {STM} & 68.7 & 78.1 & 74.4 & 46.7 & 56.4 & 52.7 & 72.0 & 81.5 & 69.2\\
    {HTM} & 68.7 & 78.2 & 74.5 & 46.1 & 55.7 & 51.5 & 72.1 & 81.2 & 69.1\\
    \midrule
    {ROMTrack (w/o vt)} & \textbf{68.8} & \textbf{78.4} & \textbf{75.0} & \textbf{48.2} & \textbf{58.4} & \textbf{54.1} & \textbf{72.5} & \textbf{82.4} & \textbf{69.8}\\
    \bottomrule
  \end{tabular}
  }
  \caption{}
  \end{subtable}
  
  \caption{ 
    (a) Architectures of OSTrack-256 (w/o CE), STM, HTM and our ROMTrack (w/o vt).
    (b) Ablation study on inherent template and hybrid template. The best results are in \textbf{bold} fonts.
    * denotes the model trained with only GOT-10k train split.
    }
    \label{tab:ablation}
\end{table*}

\begin{table}
    \centering
    \resizebox{\linewidth}{!}{
    \begin{tabular}{c|ccc|ccc}
    \toprule
    \multirow{2}{*}{Method} & \multicolumn{3}{c|}{LaSOT}  & \multicolumn{3}{c}{LaSOT$_{\textbf{ext}}$} \\
    \cline{2-7}
     & $AUC(\%)$ & $P_{Norm}(\%)$ & $P(\%)$ & $AUC(\%)$ & $P_{Norm}(\%)$ & $P(\%)$ \\
    \midrule
    {ROMTrack (w/o vt)} & 68.8 & 78.4 & 75.0 & 48.2 & 58.4 & 54.1 \\
    {ROMTrack-lpr} & 65.0 & 72.7 & 69.6 & 44.2 & 53.1 & 48.2 \\
    {ROMTrack} & \textbf{69.3} & \textbf{78.8} & \textbf{75.6} & \textbf{48.9} & \textbf{59.3} & \textbf{55.0} \\
    \midrule
    {HTM} & 68.7 & 78.2 & 74.5 & 46.1 & 55.7 & 51.5 \\
    {HTM-vt} & 69.1 & 78.6 & 75.2 & 48.6 & 58.8 & 54.5 \\
    {HTM-vt-online} & \textbf{69.6} & \textbf{79.0} & \textbf{76.0} & \textbf{49.0} & \textbf{59.4} & \textbf{55.0} \\
    \bottomrule
    \end{tabular}
    }
    \caption{Ablation study on variation tokens and exploration of template updating. The best results are in \textbf{bold} fonts.}
    \label{tab:variation_tokens_and_template_updating}
\end{table}

\noindent \textbf{Study on Variation Tokens.}
To verify the effectiveness of variation tokens, we replace the variation tokens with template features extracted using the last prediction result and form a model named ROMTrack-lpr. Specifically, suppose we get a prediction result $B_{t}$ of the target object in Frame $I_{t}$. Then in Frame $I_{t+1}$, we directly extract template features using $B_{t}$ and replace variation tokens with these feature tokens to see whether the network can automatically model appearance changes. 
As shown in the first three rows of  Table~\ref{tab:variation_tokens_and_template_updating}, ROMTrack-lpr performs even worse than ROMTrack (w/o vt) because direct employment of temporal template features tends to motivate the network to conduct more unguided feature fusion and fails to recognize appearance changes of the target object, which harms tracking performance greatly.
On the contrary, our ROMTrack outperforms ROMTrack (w/o vt) by 0.5\% and 0.7\% in AUC score on LaSOT and LaSOT$_{\textbf{ext}}$ respectively, suggesting that our variation-token design does have the capability of modeling contextual appearance changes by prompting the network to adjust appearance modeling for target objects.

\noindent \textbf{Exploration of Template Updating.}
Our method does not employ template updating, but we have also conducted some exploration studies to combine template updating with our variation-token design. 
Firstly, we add variation tokens to HTM (mentioned in Table~\ref{tab:ablation}) to form a model named HTM-vt. Secondly, based on HTM-vt, we add the online template and a simple score prediction branch borrowed from MixFormer~\cite{MixFormer} to form a model named HTM-vt-online. 
As depicted in the last three rows of Table~\ref{tab:variation_tokens_and_template_updating}, variation tokens bring significant improvement to HTM, \eg, HTM-vt outperforms HTM by 0.4\% and 2.5\% in AUC score on LaSOT and LaSOT$_{\textbf{ext}}$. 
In addition, with the online template updating strategy, HTM-vt-online further improves the AUC score by 0.5\% and 0.4\% separately. Compared to MixFormer, we take the same online strategy but achieve better performance, proving the superiority of our method.
It is concluded that our variation-token design is vital to improve the overall performance of trackers. With the assistance of a template updating strategy, the performance can be further boosted, indicating that our variation tokens work complementary with the template update strategy.

\begin{table}
    \centering
    \resizebox{\linewidth}{!}{
    \begin{tabular}{c|ccc|ccc}
    \toprule
    \multirow{2}{*}{Method} & \multicolumn{3}{c|}{LaSOT}  & \multicolumn{3}{c}{LaSOT$_{\textbf{ext}}$} \\
    \cline{2-7}
     & $AUC(\%)$ & $P_{Norm}(\%)$ & $P(\%)$ & $AUC(\%)$ & $P_{Norm}(\%)$ & $P(\%)$ \\
    \midrule
    HTM & 68.7 & 78.4 & 74.7 & 46.1 & 55.7 & 51.5 \\
    \midrule
    HTM-400 & 68.8 & 78.3 & 74.6 & 46.3 & 56.0 & 52.0 \\
    \textbf{ROMTrack} & \textbf{69.3} & \textbf{78.8} & \textbf{75.6} & \textbf{48.9} & \textbf{59.3} & \textbf{55.0} \\
    \midrule
    {ROMTrack-RS} & 69.0 & 78.4 & 75.1 & 48.4 & 58.6 & 54.3 \\
    \textbf{ROMTrack-CS} & \textbf{69.3} & \textbf{78.8} & \textbf{75.6} & \textbf{48.9} & \textbf{59.3} & \textbf{55.0} \\
    \bottomrule
    \end{tabular}
    }
    \caption{Ablation study on aligned comparison and sampling strategy. RS and CS denote random sampling and consecutive sampling. The best results are in \textbf{bold} fonts.}
    \label{tab:aligned_comparison_and_sample_strategy}
\end{table}

\noindent \textbf{Study on Aligned Comparison.}
The training of our tracker consists of two stages, adding up to 400 epochs in total. To prove that the outstanding performance of our tracker is not due to a longer training process, we select the HTM approach and train it for 400 epochs, denoted as HTM-400. The results are shown in the first two rows of Table~\ref{tab:aligned_comparison_and_sample_strategy}. Our method still outperforms HTM-400 by a large margin (\eg, +0.5\% AUC on LaSOT and +2.6\% AUC on LaSOT$_{\textbf{ext}}$), which proves that a simple extension of the training process is not the critical factor for excellent performance. Therefore, the robust object modeling approach is undoubtedly helpful for feature extraction and relation modeling.

\noindent \textbf{Study on Sampling Strategy.}
During the second training stage, we employ a particular sampling strategy called consecutive sampling (CS). 
Different from random sampling (RS), two consecutive frames instead of two random frames in the same video sequence are sampled as the search region. The training process is described in Section~\ref{sec:discussions}. 
The results in Table~\ref{tab:aligned_comparison_and_sample_strategy} show that the consecutive sampling strategy obtains better performance because it helps the model to learn more about the temporal variation of object appearance. In addition, the model trained with random sampling also performs better than HTM, indicating the effectiveness of our robust object modeling.




\section{Conclusions}
This work proposes a robust object modeling framework for visual tracking (ROMTrack). 
The proposed ROMTrack utilizes two template streams to learn robust and discriminative feature representations.
The inherent template keeps original features of target objects, and the hybrid template learns mixed template-search features. The hybrid template can extract helpful information from the inherent template to form target-oriented features.
Besides, the variation tokens are introduced to embed appearance context, thus adaptable to object deformation and appearance variations. The variation-token design is subtly integrated into attention computation, leading to a neat and effective tracker. 
Extensive experiments show that ROMTrack performs better than previous methods on multiple benchmarks. 

{\small
\bibliographystyle{ieee_fullname}
\bibliography{egbib}

\begin{thebibliography}{10}\itemsep=-1pt

\bibitem{Staple}
Luca Bertinetto, Jack Valmadre, Stuart Golodetz, Ondrej Miksik, and Philip
  H.~S. Torr.
\newblock Staple: Complementary learners for real-time tracking.
\newblock In {\em {CVPR}}, pages 1401--1409. {IEEE} Computer Society, 2016.

\bibitem{SiamFC}
Luca Bertinetto, Jack Valmadre, Jo{\~{a}}o~F. Henriques, Andrea Vedaldi, and
  Philip H.~S. Torr.
\newblock Fully-convolutional siamese networks for object tracking.
\newblock In {\em {ECCV} Workshops {(2)}}, volume 9914 of {\em Lecture Notes in
  Computer Science}, pages 850--865, 2016.

\bibitem{DiMP}
Goutam Bhat, Martin Danelljan, Luc~Van Gool, and Radu Timofte.
\newblock Learning discriminative model prediction for tracking.
\newblock In {\em {ICCV}}, pages 6181--6190. {IEEE}, 2019.

\bibitem{BhatJDKF18}
Goutam Bhat, Joakim Johnander, Martin Danelljan, Fahad~Shahbaz Khan, and
  Michael Felsberg.
\newblock Unveiling the power of deep tracking.
\newblock In {\em {ECCV} {(2)}}, volume 11206 of {\em Lecture Notes in Computer
  Science}, pages 493--509. Springer, 2018.

\bibitem{UPDT}
Goutam Bhat, Joakim Johnander, Martin Danelljan, Fahad~Shahbaz Khan, and
  Michael Felsberg.
\newblock Unveiling the power of deep tracking.
\newblock In {\em {ECCV} {(2)}}, volume 11206 of {\em Lecture Notes in Computer
  Science}, pages 493--509. Springer, 2018.

\bibitem{SimTrack}
Boyu Chen, Peixia Li, Lei Bai, Lei Qiao, Qiuhong Shen, Bo Li, Weihao Gan, Wei
  Wu, and Wanli Ouyang.
\newblock Backbone is all your need: {A} simplified architecture for visual
  object tracking.
\newblock In {\em {ECCV} {(22)}}, volume 13682 of {\em Lecture Notes in
  Computer Science}, pages 375--392. Springer, 2022.

\bibitem{TransT}
Xin Chen, Bin Yan, Jiawen Zhu, Dong Wang, Xiaoyun Yang, and Huchuan Lu.
\newblock Transformer tracking.
\newblock In {\em {CVPR}}, pages 8126--8135. Computer Vision Foundation /
  {IEEE}, 2021.

\bibitem{SiamBAN}
Zedu Chen, Bineng Zhong, Guorong Li, Shengping Zhang, and Rongrong Ji.
\newblock Siamese box adaptive network for visual tracking.
\newblock In {\em {CVPR}}, pages 6667--6676. Computer Vision Foundation /
  {IEEE}, 2020.

\bibitem{FCOT}
Yutao Cui, Cheng Jiang, Limin Wang, and Gangshan Wu.
\newblock Fully convolutional online tracking.
\newblock {\em CoRR}, abs/2004.07109, 2020.

\bibitem{TREG}
Yutao Cui, Cheng Jiang, Limin Wang, and Gangshan Wu.
\newblock Target transformed regression for accurate tracking.
\newblock {\em CoRR}, abs/2104.00403, 2021.

\bibitem{MixFormer}
Yutao Cui, Cheng Jiang, Limin Wang, and Gangshan Wu.
\newblock Mixformer: End-to-end tracking with iterative mixed attention.
\newblock In {\em {CVPR}}, pages 13598--13608. {IEEE}, 2022.

\bibitem{LTMU}
Kenan Dai, Yunhua Zhang, Dong Wang, Jianhua Li, Huchuan Lu, and Xiaoyun Yang.
\newblock High-performance long-term tracking with meta-updater.
\newblock In {\em {CVPR}}, pages 6297--6306. Computer Vision Foundation /
  {IEEE}, 2020.

\bibitem{ATOM}
Martin Danelljan, Goutam Bhat, Fahad~Shahbaz Khan, and Michael Felsberg.
\newblock {ATOM:} accurate tracking by overlap maximization.
\newblock In {\em {CVPR}}, pages 4660--4669. Computer Vision Foundation /
  {IEEE}, 2019.

\bibitem{PrDiMP}
Martin Danelljan, Luc~Van Gool, and Radu Timofte.
\newblock Probabilistic regression for visual tracking.
\newblock In {\em {CVPR}}, pages 7181--7190. Computer Vision Foundation /
  {IEEE}, 2020.

\bibitem{ImageNet}
Jia Deng, Wei Dong, Richard Socher, Li{-}Jia Li, Kai Li, and Li Fei{-}Fei.
\newblock Imagenet: {A} large-scale hierarchical image database.
\newblock In {\em {CVPR}}, pages 248--255. {IEEE} Computer Society, 2009.

\bibitem{ViT}
Alexey Dosovitskiy, Lucas Beyer, Alexander Kolesnikov, Dirk Weissenborn,
  Xiaohua Zhai, Thomas Unterthiner, Mostafa Dehghani, Matthias Minderer, Georg
  Heigold, Sylvain Gelly, Jakob Uszkoreit, and Neil Houlsby.
\newblock An image is worth 16x16 words: Transformers for image recognition at
  scale.
\newblock In {\em {ICLR}}. OpenReview.net, 2021.

\bibitem{CGACD}
Fei Du, Peng Liu, Wei Zhao, and Xianglong Tang.
\newblock Correlation-guided attention for corner detection based visual
  tracking.
\newblock In {\em {CVPR}}, pages 6835--6844. Computer Vision Foundation /
  {IEEE}, 2020.

\bibitem{LaSOT_ext}
Heng Fan, Hexin Bai, Liting Lin, Fan Yang, Peng Chu, Ge Deng, Sijia Yu,
  Harshit, Mingzhen Huang, Juehuan Liu, Yong Xu, Chunyuan Liao, Lin Yuan, and
  Haibin Ling.
\newblock Lasot: {A} high-quality large-scale single object tracking benchmark.
\newblock {\em Int. J. Comput. Vis.}, 129(2):439--461, 2021.

\bibitem{LaSOT}
Heng Fan, Liting Lin, Fan Yang, Peng Chu, Ge Deng, Sijia Yu, Hexin Bai, Yong
  Xu, Chunyuan Liao, and Haibin Ling.
\newblock Lasot: {A} high-quality benchmark for large-scale single object
  tracking.
\newblock In {\em {CVPR}}, pages 5374--5383. Computer Vision Foundation /
  {IEEE}, 2019.

\bibitem{CRPN}
Heng Fan and Haibin Ling.
\newblock Siamese cascaded region proposal networks for real-time visual
  tracking.
\newblock In {\em {CVPR}}, pages 7952--7961. Computer Vision Foundation /
  {IEEE}, 2019.

\bibitem{STMTrack}
Zhihong Fu, Qingjie Liu, Zehua Fu, and Yunhong Wang.
\newblock Stmtrack: Template-free visual tracking with space-time memory
  networks.
\newblock In {\em {CVPR}}, pages 13774--13783. Computer Vision Foundation /
  {IEEE}, 2021.

\bibitem{NFS30}
Hamed~Kiani Galoogahi, Ashton Fagg, Chen Huang, Deva Ramanan, and Simon Lucey.
\newblock Need for speed: {A} benchmark for higher frame rate object tracking.
\newblock In {\em {ICCV}}, pages 1134--1143. {IEEE} Computer Society, 2017.

\bibitem{GaloogahiFL17}
Hamed~Kiani Galoogahi, Ashton Fagg, and Simon Lucey.
\newblock Learning background-aware correlation filters for visual tracking.
\newblock In {\em {ICCV}}, pages 1144--1152. {IEEE} Computer Society, 2017.

\bibitem{AiATrack}
Shenyuan Gao, Chunluan Zhou, Chao Ma, Xinggang Wang, and Junsong Yuan.
\newblock Aiatrack: Attention in attention for transformer visual tracking.
\newblock In {\em {ECCV} {(22)}}, volume 13682 of {\em Lecture Notes in
  Computer Science}, pages 146--164. Springer, 2022.

\bibitem{MAE}
Kaiming He, Xinlei Chen, Saining Xie, Yanghao Li, Piotr Doll{\'{a}}r, and
  Ross~B. Girshick.
\newblock Masked autoencoders are scalable vision learners.
\newblock In {\em {CVPR}}, pages 15979--15988. {IEEE}, 2022.

\bibitem{KCF}
Jo{\~{a}}o~F. Henriques, Rui Caseiro, Pedro Martins, and Jorge Batista.
\newblock High-speed tracking with kernelized correlation filters.
\newblock {\em {IEEE} Trans. Pattern Anal. Mach. Intell.}, 37(3):583--596,
  2015.

\bibitem{GOT-10k}
Lianghua Huang, Xin Zhao, and Kaiqi Huang.
\newblock Got-10k: {A} large high-diversity benchmark for generic object
  tracking in the wild.
\newblock {\em {IEEE} Trans. Pattern Anal. Mach. Intell.}, 43(5):1562--1577,
  2021.

\bibitem{VOT2018}
Matej Kristan, Ales Leonardis, et~al.
\newblock The sixth visual object tracking {VOT2018} challenge results.
\newblock In {\em {ECCV} Workshops {(1)}}, volume 11129 of {\em Lecture Notes
  in Computer Science}, pages 3--53. Springer, 2018.

\bibitem{VOT2020}
Matej Kristan, Ales Leonardis, et~al.
\newblock The eighth visual object tracking {VOT2020} challenge results.
\newblock In {\em {ECCV} Workshops {(5)}}, volume 12539 of {\em Lecture Notes
  in Computer Science}, pages 547--601. Springer, 2020.

\bibitem{CornerNet}
Hei Law and Jia Deng.
\newblock Cornernet: Detecting objects as paired keypoints.
\newblock {\em Int. J. Comput. Vis.}, 128(3):642--656, 2020.

\bibitem{SiamRPN++}
Bo Li, Wei Wu, Qiang Wang, Fangyi Zhang, Junliang Xing, and Junjie Yan.
\newblock Siamrpn++: Evolution of siamese visual tracking with very deep
  networks.
\newblock In {\em {CVPR}}, pages 4282--4291. Computer Vision Foundation /
  {IEEE}, 2019.

\bibitem{SiameseRPN}
Bo Li, Junjie Yan, Wei Wu, Zheng Zhu, and Xiaolin Hu.
\newblock High performance visual tracking with siamese region proposal
  network.
\newblock In {\em {CVPR}}, pages 8971--8980. Computer Vision Foundation /
  {IEEE} Computer Society, 2018.

\bibitem{SwinTrack}
Liting Lin, Heng Fan, Yong Xu, and Haibin Ling.
\newblock Swintrack: {A} simple and strong baseline for transformer tracking.
\newblock {\em CoRR}, abs/2112.00995, 2021.

\bibitem{COCO}
Tsung{-}Yi Lin, Michael Maire, Serge~J. Belongie, James Hays, Pietro Perona,
  Deva Ramanan, Piotr Doll{\'{a}}r, and C.~Lawrence Zitnick.
\newblock Microsoft {COCO:} common objects in context.
\newblock In {\em {ECCV} {(5)}}, volume 8693 of {\em Lecture Notes in Computer
  Science}, pages 740--755. Springer, 2014.

\bibitem{SwinTransformer}
Ze Liu, Yutong Lin, Yue Cao, Han Hu, Yixuan Wei, Zheng Zhang, Stephen Lin, and
  Baining Guo.
\newblock Swin transformer: Hierarchical vision transformer using shifted
  windows.
\newblock In {\em {ICCV}}, pages 9992--10002. {IEEE}, 2021.

\bibitem{AdamW}
Ilya Loshchilov and Frank Hutter.
\newblock Decoupled weight decay regularization.
\newblock In {\em {ICLR} (Poster)}. OpenReview.net, 2019.

\bibitem{LukezicVZMK17}
Alan Lukezic, Tomas Vojir, Luka~Cehovin Zajc, Jiri Matas, and Matej Kristan.
\newblock Discriminative correlation filter with channel and spatial
  reliability.
\newblock In {\em {CVPR}}, pages 4847--4856. {IEEE} Computer Society, 2017.

\bibitem{ToMP}
Christoph Mayer, Martin Danelljan, Goutam Bhat, Matthieu Paul, Danda~Pani
  Paudel, Fisher Yu, and Luc~Van Gool.
\newblock Transforming model prediction for tracking.
\newblock In {\em {CVPR}}, pages 8721--8730. {IEEE}, 2022.

\bibitem{KeepTrack}
Christoph Mayer, Martin Danelljan, Danda~Pani Paudel, and Luc~Van Gool.
\newblock Learning target candidate association to keep track of what not to
  track.
\newblock In {\em {ICCV}}, pages 13424--13434. {IEEE}, 2021.

\bibitem{UAV123}
Matthias Mueller, Neil Smith, and Bernard Ghanem.
\newblock A benchmark and simulator for {UAV} tracking.
\newblock In {\em {ECCV} {(1)}}, volume 9905 of {\em Lecture Notes in Computer
  Science}, pages 445--461. Springer, 2016.

\bibitem{TrackingNet}
Matthias M{\"{u}}ller, Adel Bibi, Silvio Giancola, Salman Al{-}Subaihi, and
  Bernard Ghanem.
\newblock Trackingnet: {A} large-scale dataset and benchmark for object
  tracking in the wild.
\newblock In {\em {ECCV} {(1)}}, volume 11205 of {\em Lecture Notes in Computer
  Science}, pages 310--327. Springer, 2018.

\bibitem{MDNet}
Hyeonseob Nam and Bohyung Han.
\newblock Learning multi-domain convolutional neural networks for visual
  tracking.
\newblock In {\em {CVPR}}, pages 4293--4302. {IEEE} Computer Society, 2016.

\bibitem{RTS}
Matthieu Paul, Martin Danelljan, Christoph Mayer, and Luc~Van Gool.
\newblock Robust visual tracking by segmentation.
\newblock In {\em {ECCV} {(22)}}, volume 13682 of {\em Lecture Notes in
  Computer Science}, pages 571--588. Springer, 2022.

\bibitem{Song0WGBZSL018}
Yibing Song, Chao Ma, Xiaohe Wu, Lijun Gong, Linchao Bao, Wangmeng Zuo, Chunhua
  Shen, Rynson W.~H. Lau, and Ming{-}Hsuan Yang.
\newblock {VITAL:} visual tracking via adversarial learning.
\newblock In {\em {CVPR}}, pages 8990--8999. Computer Vision Foundation /
  {IEEE} Computer Society, 2018.

\bibitem{AttentionIsAllYouNeed}
Ashish Vaswani, Noam Shazeer, Niki Parmar, Jakob Uszkoreit, Llion Jones,
  Aidan~N. Gomez, Lukasz Kaiser, and Illia Polosukhin.
\newblock Attention is all you need.
\newblock In {\em {NIPS}}, pages 5998--6008, 2017.

\bibitem{SiamR-CNN}
Paul Voigtlaender, Jonathon Luiten, Philip H.~S. Torr, and Bastian Leibe.
\newblock Siam {R-CNN:} visual tracking by re-detection.
\newblock In {\em {CVPR}}, pages 6577--6587. Computer Vision Foundation /
  {IEEE}, 2020.

\bibitem{TrDiMP}
Ning Wang, Wengang Zhou, Jie Wang, and Houqiang Li.
\newblock Transformer meets tracker: Exploiting temporal context for robust
  visual tracking.
\newblock In {\em {CVPR}}, pages 1571--1580. Computer Vision Foundation /
  {IEEE}, 2021.

\bibitem{RASNet}
Qiang Wang, Zhu Teng, Junliang Xing, Jin Gao, Weiming Hu, and Stephen~J.
  Maybank.
\newblock Learning attentions: Residual attentional siamese network for high
  performance online visual tracking.
\newblock In {\em {CVPR}}, pages 4854--4863. Computer Vision Foundation /
  {IEEE} Computer Society, 2018.

\bibitem{PVT}
Wenhai Wang, Enze Xie, Xiang Li, Deng{-}Ping Fan, Kaitao Song, Ding Liang, Tong
  Lu, Ping Luo, and Ling Shao.
\newblock Pyramid vision transformer: {A} versatile backbone for dense
  prediction without convolutions.
\newblock In {\em {ICCV}}, pages 548--558. {IEEE}, 2021.

\bibitem{WangZLWTB21}
Zhongdao Wang, Hengshuang Zhao, Ya{-}Li Li, Shengjin Wang, Philip H.~S. Torr,
  and Luca Bertinetto.
\newblock Do different tracking tasks require different appearance models?
\newblock In {\em NeurIPS}, pages 726--738, 2021.

\bibitem{CvT}
Haiping Wu, Bin Xiao, Noel Codella, Mengchen Liu, Xiyang Dai, Lu Yuan, and Lei
  Zhang.
\newblock Cvt: Introducing convolutions to vision transformers.
\newblock In {\em {ICCV}}, pages 22--31. {IEEE}, 2021.

\bibitem{OTB100}
Yi Wu, Jongwoo Lim, and Ming{-}Hsuan Yang.
\newblock Object tracking benchmark.
\newblock {\em {IEEE} Trans. Pattern Anal. Mach. Intell.}, 37(9):1834--1848,
  2015.

\bibitem{SBT}
Fei Xie, Chunyu Wang, Guangting Wang, Yue Cao, Wankou Yang, and Wenjun Zeng.
\newblock Correlation-aware deep tracking.
\newblock In {\em {CVPR}}, pages 8741--8750. {IEEE}, 2022.

\bibitem{SiamFC++}
Yinda Xu, Zeyu Wang, Zuoxin Li, Yuan Ye, and Gang Yu.
\newblock Siamfc++: Towards robust and accurate visual tracking with target
  estimation guidelines.
\newblock In {\em {AAAI}}, pages 12549--12556. {AAAI} Press, 2020.

\bibitem{Unicorn}
Bin Yan, Yi Jiang, Peize Sun, Dong Wang, Zehuan Yuan, Ping Luo, and Huchuan Lu.
\newblock Towards grand unification of object tracking.
\newblock In {\em {ECCV} {(21)}}, volume 13681 of {\em Lecture Notes in
  Computer Science}, pages 733--751. Springer, 2022.

\bibitem{STARK}
Bin Yan, Houwen Peng, Jianlong Fu, Dong Wang, and Huchuan Lu.
\newblock Learning spatio-temporal transformer for visual tracking.
\newblock In {\em {ICCV}}, pages 10428--10437. {IEEE}, 2021.

\bibitem{OSTrack}
Botao Ye, Hong Chang, Bingpeng Ma, Shiguang Shan, and Xilin Chen.
\newblock Joint feature learning and relation modeling for tracking: {A}
  one-stream framework.
\newblock In {\em {ECCV} {(22)}}, volume 13682 of {\em Lecture Notes in
  Computer Science}, pages 341--357. Springer, 2022.

\bibitem{DTT}
Bin Yu, Ming Tang, Linyu Zheng, Guibo Zhu, Jinqiao Wang, Hao Feng, Xuetao Feng,
  and Hanqing Lu.
\newblock High-performance discriminative tracking with transformers.
\newblock In {\em {ICCV}}, pages 9836--9845. {IEEE}, 2021.

\bibitem{SiamAttn}
Yuechen Yu, Yilei Xiong, Weilin Huang, and Matthew~R. Scott.
\newblock Deformable siamese attention networks for visual object tracking.
\newblock In {\em {CVPR}}, pages 6727--6736. Computer Vision Foundation /
  {IEEE}, 2020.

\bibitem{UpdateNet}
Lichao Zhang, Abel Gonzalez{-}Garcia, Joost van~de Weijer, Martin Danelljan,
  and Fahad~Shahbaz Khan.
\newblock Learning the model update for siamese trackers.
\newblock In {\em {ICCV}}, pages 4009--4018. {IEEE}, 2019.

\bibitem{Ocean}
Zhipeng Zhang, Houwen Peng, Jianlong Fu, Bing Li, and Weiming Hu.
\newblock Ocean: Object-aware anchor-free tracking.
\newblock In {\em {ECCV} {(21)}}, volume 12366 of {\em Lecture Notes in
  Computer Science}, pages 771--787. Springer, 2020.

\bibitem{CenterNet}
Xingyi Zhou, Dequan Wang, and Philipp Kr{\"{a}}henb{\"{u}}hl.
\newblock Objects as points.
\newblock {\em CoRR}, abs/1904.07850, 2019.

\bibitem{DaSiamRPN}
Zheng Zhu, Qiang Wang, Bo Li, Wei Wu, Junjie Yan, and Weiming Hu.
\newblock Distractor-aware siamese networks for visual object tracking.
\newblock In {\em {ECCV} {(9)}}, volume 11213 of {\em Lecture Notes in Computer
  Science}, pages 103--119. Springer, 2018.

\end{thebibliography}
}

\clearpage

\section*{Appendix}
The supplementary material provides more details of our implementation in Section~\ref{sec:ExpDetails}, more results on other datasets (UAV123~\cite{UAV123} and VOT2020~\cite{VOT2020}) in Section~\ref{sec:MoreResults}, more analysis on LaSOT~\cite{LaSOT} and LaSOT$_{\textbf{ext}}$~\cite{LaSOT_ext} in Section~\ref{sec:MoreAnalysis}, and visualization results in Section~\ref{sec:Visualizations}.

\appendix
\section{Experiment Details}
\label{sec:ExpDetails}

\subsection{Training Details}
\noindent \textbf{Pre-trained Model and Architecture Details.} We adopt the vanilla ViT-Base~\cite{ViT} model pre-trained with MAE~\cite{MAE} on ImageNet-1k to initialize the backbone of our ROMTrack. More specifically, the original input of the pre-trained model is $224\times 224$ in resolution. In contrast, the resolution of the template and search images of our ROMTrack are $128\times 128$ pixels and $256\times 256$ pixels, respectively. And for our ROMTrack-384, the resolution are $192\times 192$ and $384\times 384$. So we employ the bicubic interpolation to generate appropriate position embeddings for the template tokens and the search region tokens separately. Both template and search images share the rest weights of the backbone. 

The prediction head of our ROMTrack is a center-based head consisting of 4 stacked Conv-BN-ReLU layers for each output. We initialize the weights of the head with a Xavier uniform initializer.

\noindent \textbf{Experimental Settings.} The whole training process of our ROMTrack consists of two stages: the first 300 epochs are to train the backbone and the head without variation tokens, and the extra 100 epochs are to merge the variation tokens into the architecture. 

In the first stage, we randomly sample a $256\times 256$ search region plus two $128\times 128$ template regions in the same sequence as input. One of the templates serves as the hybrid template to participate in the calculation of hybrid features. Another template is the inherent template to extract inherent features layer by layer. The training recipe is similar to conventional trackers.
In the second stage, the variation tokens are merged into the architecture. We employ a particular sampling strategy called consecutive sampling (CS). Namely, we consecutively sample two $256\times 256$ search regions and randomly sample two $128\times 128$ template regions as input. These four images are still in the same sequence. We use the first search region and two template regions to conduct tracking similar to the first stage, and the hybrid template features are preserved as the variation tokens. Then we use the second search region and another two template regions together with the preserved variation tokens for robust tracking training.

Generally, we use 4 Tesla V100 GPUs to train ROMTrack with a batch size of 64 per GPU. For ROMTrack-384, we use 8 Tesla V100 GPUs with a batch size of 16 per GPU due to increased GPU memory consumption. We apply the gradient clip strategy with a clip normalization rate of 0.1. Both training stages have 60k image pairs per epoch. The GOT-10k test benchmark requires a one-shot setting, so we set the training epoch of the first stage to 100 and decrease the learning rate by a factor of 10 after 80 epochs. Similarly, we set the training epoch of the second stage to 50 and decrease the learning rate by a factor of 10 after 40 epochs. 

\noindent \textbf{Ablation Settings.} To make a fair comparison, the methods (\eg, STM and HTM) mentioned in the ablation study are all implemented under the same framework and experimental settings as our ROMTrack. Namely, the pre-trained model, batch size, training pairs, and learning rate strategy are all kept consistent with our ROMTrack.
Besides, in the exploration studies of template updating, the online branch we take is the SPM module introduced by MixFormer~\cite{MixFormer}. The borrowed SPM module enhances our tracker's performance and further widens the performance gap with MixFormer, proving the superiority of our method.

\subsection{Inference Details}
Unlike the training process, we directly employ the initial template as both the hybrid template and the inherent template during inference. The variation tokens are obtained by preserving the hybrid template features per frame. In this way, the inherent feature of the initial template is enhanced, and the contextual appearance changes captured by variation tokens can be utilized to facilitate the tracking procedure.


\begin{table*}
  \centering
  \resizebox{\linewidth}{!}{
  \begin{tabular}{c|ccccccccccc|cc}
    \toprule
    & SiamRPN++ & PrDiMP & TransT & STARK & KeepTrack & ToMP & MixFormer & OSTrack & OSTrack & SwinTrack & SwinTrack & ROMTrack & ROMTrack-384 \\
    & ~\cite{SiamRPN++} & ~\cite{PrDiMP} & ~\cite{TransT} & ~\cite{STARK} & ~\cite{KeepTrack} & 101~\cite{ToMP} & L~\cite{MixFormer} & 256~\cite{OSTrack} & 384~\cite{OSTrack} & T-224~\cite{SwinTrack} & B-384~\cite{SwinTrack} & (Ours) & (Ours) \\
    \midrule
    AUC($\%$) & 61.3 & 68.0 & 69.1 & 69.1 & 69.7 & 66.9 & 69.5 & 68.3 & \color{red}\textbf{70.7} & 68.8 & 70.5 & 69.7 & \color{blue}\textbf{70.5} \\
    \bottomrule
  \end{tabular}
  }
  \caption{Comparison with state-of-the-art trackers on UAV123. The best two results are shown in \textcolor{red}{\textbf{red}} and \textcolor{blue}{\textbf{blue}} fonts.}
  \label{tab:uav}
\end{table*}

\begin{table*}
  \centering
  \resizebox{\linewidth}{!}{
  \begin{tabular}{c|cccccccccccc|cc}
    \toprule
     & SiamFC & ATOM & DiMP & UPDT & TRAT & SuperDiMP & STARK & STARK & ToMP & ToMP & SwinTrack & SwinTrack & ROMTrack & ROMTrack-384 \\
     & ~\cite{SiamFC} & ~\cite{ATOM} & ~\cite{DiMP} & ~\cite{UPDT} & ~\cite{VOT2020} & ~\cite{VOT2018} & ST50~\cite{STARK} & ST101~\cite{STARK} & 50~\cite{ToMP} & 101~\cite{ToMP} & T-224~\cite{SwinTrack} & B-384~\cite{SwinTrack} & (Ours) & (Ours) \\
    \midrule
    EAO & 0.179 & 0.271 & 0.274 & 0.278 & 0.280 & 0.305 & 0.308 & 0.303 & 0.297 & 0.309 & 0.302 & 0.283 & \color{blue}\textbf{0.326} & \color{red}\textbf{0.329} \\
    Accuracy & 0.418 & 0.462 & 0.457 & 0.465 & 0.464 & 0.477 & 0.478 & \color{blue}\textbf{0.481} & 0.453 & 0.453 & 0.471 & 0.472 & 0.480 & \color{red}\textbf{0.483} \\
    Robustness & 0.502 & 0.734 & 0.740 & 0.755 & 0.744 & 0.786 & 0.799 & 0.775 & 0.789 & 0.814 & 0.775 & 0.741 & \color{blue}\textbf{0.816} & \color{red}\textbf{0.822} \\
    \bottomrule
  \end{tabular}
  }
  \caption{Comparison with state-of-the-art trackers (bounding box only methods) on VOT2020. The best two results are shown in \textcolor{red}{\textbf{red}} and \textcolor{blue}{\textbf{blue}}.}
  \label{tab:vot2020}
\end{table*}

\section{More Results}
\label{sec:MoreResults}
\subsection{Results on UAV123 Benchmark}
UAV123~\cite{UAV123} is an aerial video dataset captured from low-altitude UAVs, which contains 123 sequences with an average sequence length of 915 frames.

The results in Table~\ref{tab:uav} show that our ROMTrack outperforms OSTrack-256 and SwinTrack-T-224 appreciably. Meanwhile, our ROMTrack-384 achieves competitive performance with previous SOTA trackers.

\subsection{Results on VOT2020 Benchmark}
VOT2020~\cite{VOT2020} consists of 60 videos with segmentation masks annotated. It concentrates on several challenges, including fast motion and occlusion. Since our design is a bounding-box-only method, we also compare our results with trackers that predict the bounding boxes. 

Table~\ref{tab:vot2020} shows that ROMTrack-384 achieves the top-ranked performance on an EAO score of 0.329, surpassing the previous SOTA tracker ToMP101 with a large margin of 2\%.
Besides, both ROMTrack and ROMTrack-384 show excellent performance on the Robustness score, suggesting that our robust object modeling method significantly improves the robustness of the tracker and finally boosts the overall performance.

\section{More Analysis}
\label{sec:MoreAnalysis}

\begin{figure*}
    \centering
    \includegraphics[page=1, width=.33\linewidth]{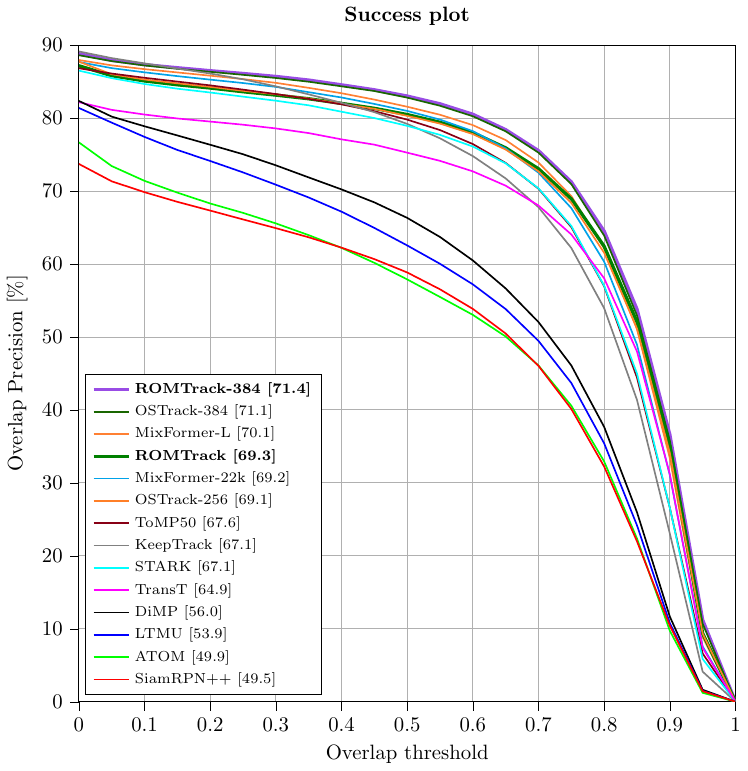}
    \includegraphics[page=2, width=.33\linewidth]{Figures/plot.pdf}
    \includegraphics[page=3, width=.33\linewidth]{Figures/plot.pdf}
    \caption{Success plot, Precision plot, and Normalized Precision plot for LaSOT.}
    \label{fig:lasot_plot}
\end{figure*}

\begin{figure*}
    \centering
    \includegraphics[page=4, width=.33\linewidth]{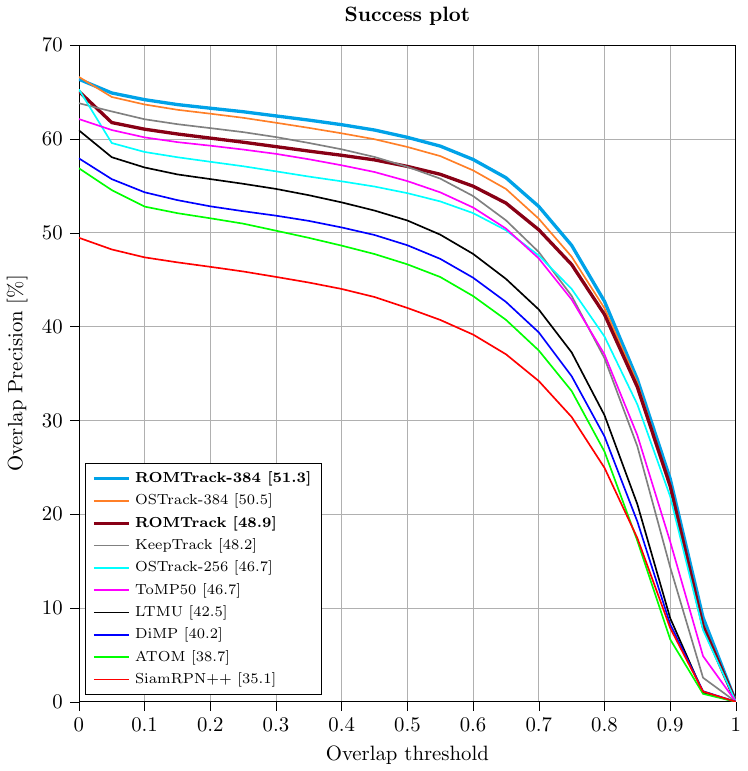}
    \includegraphics[page=5, width=.33\linewidth]{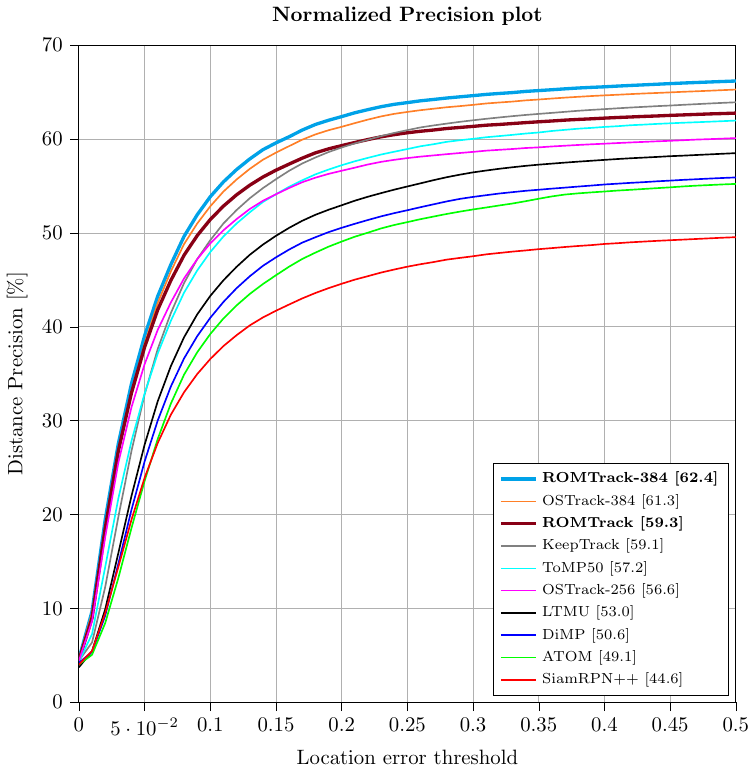}
    \includegraphics[page=6, width=.33\linewidth]{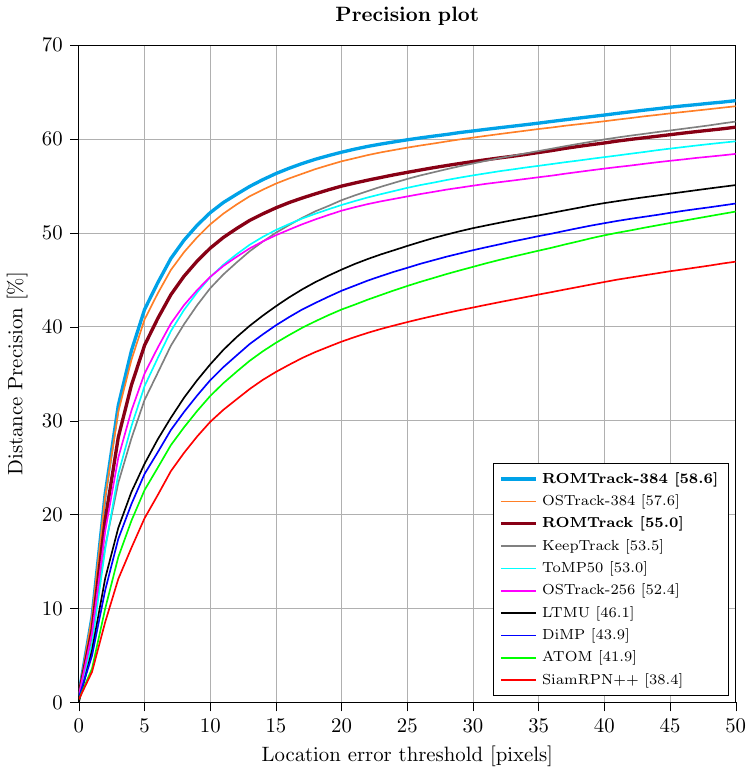}
    \caption{Success plot, Precision plot, and Normalized Precision plot for LaSOT$_{\textbf{ext}}$.}
    \label{fig:lasotext_plot}
\end{figure*}

\begin{figure*}
  \centering
  \includegraphics[width=.9\linewidth]{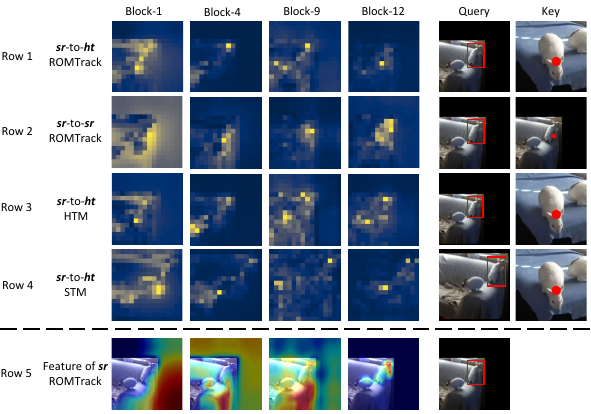}

   \caption{Visualizations of attention maps and features after different blocks. The \textcolor{red}{red} rectangle is the ground-truth box which indicates the target object to be tracked. $\textbf{sr}$-to-$\textbf{ht}$ refers to search-to-template attention and $\textbf{sr}$-to-$\textbf{sr}$ refers to search-to-search attention. The \textcolor{red}{red} circle stands for the part chosen as the key, and the attention maps are generated with the query and key pair. It can be observed that our method effectively concentrates on the target object rather than the distractor.}
   \label{fig:vis}
\end{figure*}

\noindent \textbf{LaSOT.} LaSOT~\cite{LaSOT} test set contains 280 video sequences for long-term tracking. We present the Success plot, Precision plot, and Normalized Precision plot for LaSOT in Figure~\ref{fig:lasot_plot} to conduct further comparison and analysis.

\noindent \textbf{LaSOT$_{\textbf{ext}}$.} LaSOT$_{\textbf{ext}}$~\cite{LaSOT_ext} contains 150 video sequences of 15 new object classes for long-term tracking. We present the Success plot, Precision plot, and Normalized Precision plot for LaSOT$_{\textbf{ext}}$ in Figure~\ref{fig:lasotext_plot} to conduct further comparison and analysis.

\noindent \textbf{Discussion.} 
Although an improvement on the LaSOT dataset is difficult due to the long-term attribute, we can continue to considerably boost the overall performance (+0.6\% AUC compared to OSTrack-256 (w/o CE)) by using our robust object modeling, even without a template updating strategy.
All the plots presented above prove that our ROMTrack and ROMTrack-384 improve the tracking results in both accuracy and robustness, showing superior performances of our methods.

\section{Visualizations}
\label{sec:Visualizations}
In this section, we first make a comparison between our ROMTrack and the methods proposed in ablation (\ie, HTM and STM) based on visualization results.
Then we provide more visualization results of attention maps and feature maps on LaSOT. 

\subsection{Visualization Results}
To explore how the robust object modeling works in our framework, we also visualize some attention maps and search region features in Figure~\ref{fig:vis}.
We observe that:
\begin{itemize} 
   \item the object in the search region is enhanced layer by layer through interaction with the two template streams and the variation tokens.
   \item possible distractors in the background get suppressed with our ROMTrack (Row 1, Row 2, and Row 5), suggesting the robustness of our method.
   \item both HTM (Row 3) and STM (Row 4) have difficulty in distinguishing distractors with target objects while our ROMTrack locates objects more accurately.
\end{itemize}

It can be observed that our ROMTrack concentrates on the target object rather than the distractor, showing good accuracy and robustness. As a result, our ROMTrack shows excellent tracking performance.

\subsection{More Visualizations}
More visualization results are presented in this section.
These results indicate that the proposed robust object modeling method works well under various target categories and challenging scenarios.

\begin{figure*}
    \centering
    \includegraphics[page=1, width=.795\linewidth]{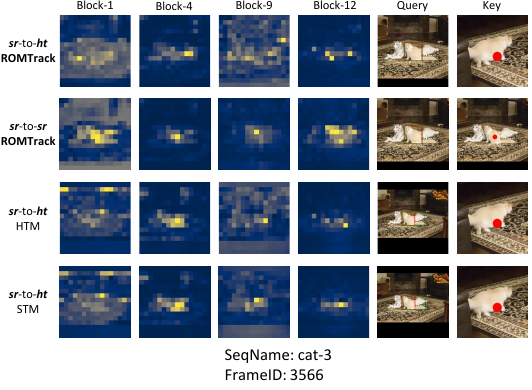}
    \includegraphics[page=3, width=.795\linewidth]{Figures/vis_supple.pdf}
    \caption{Visualizations of attention maps after different blocks. The \textcolor{red}{red} rectangle is the ground-truth box which indicates the target object to be tracked. 
    The \textcolor{green}{green} rectangle indicates the predicted bounding box of the target object.
    $\textbf{sr}$-to-$\textbf{ht}$ refers to search-to-template attention and $\textbf{sr}$-to-$\textbf{sr}$ refers to search-to-search attention. The \textcolor{red}{red} circle stands for the part chosen as the key, and the attention maps are generated with the query and key pair.
    Best viewed by zooming in.}
    \label{fig:attn_vis1}
\end{figure*}

\begin{figure*}
    \centering
    \includegraphics[page=5, width=.795\linewidth]{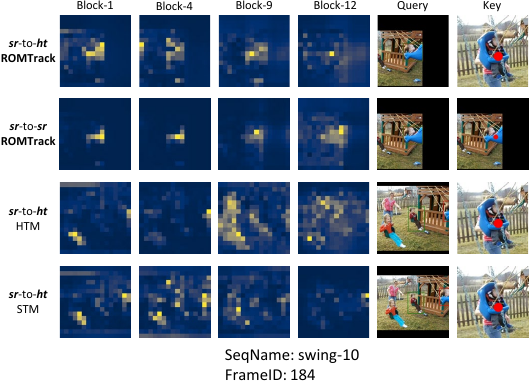}
    \includegraphics[page=7, width=.795\linewidth]{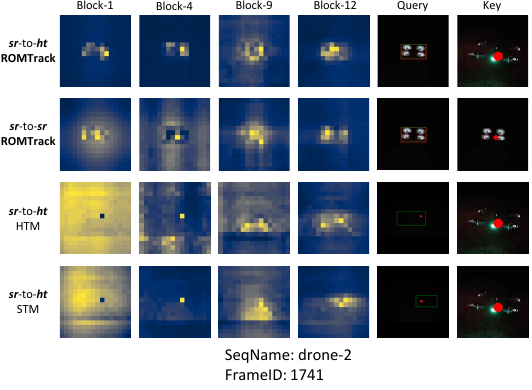}
    \caption{Visualizations of attention maps after different blocks. The \textcolor{red}{red} rectangle is the ground-truth box which indicates the target object to be tracked. 
    The \textcolor{green}{green} rectangle indicates the predicted bounding box of the target object.
    $\textbf{sr}$-to-$\textbf{ht}$ refers to search-to-template attention and $\textbf{sr}$-to-$\textbf{sr}$ refers to search-to-search attention. The \textcolor{red}{red} circle stands for the part chosen as the key, and the attention maps are generated with the query and key pair.
    Best viewed by zooming in.}
    \label{fig:attn_vis2}
\end{figure*}

\noindent \textbf{Attention Maps.} Figure~\ref{fig:attn_vis1} and Figure~\ref{fig:attn_vis2} present more attention maps of different methods. By comparing the $sr$-to-$ht$ attention maps, we can find that our ROMTrack effectively concentrates on the target object rather than the distractor, showing a more robust performance than HTM and STM.

\begin{figure*}
    \centering
    \includegraphics[page=2, width=.985\linewidth]{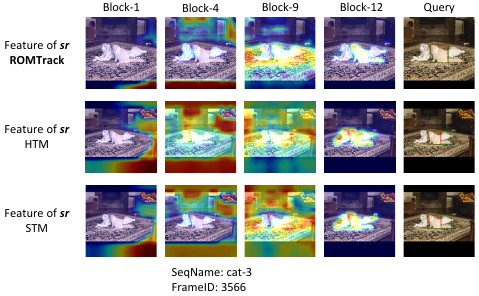}
    \includegraphics[page=4, width=.985\linewidth]{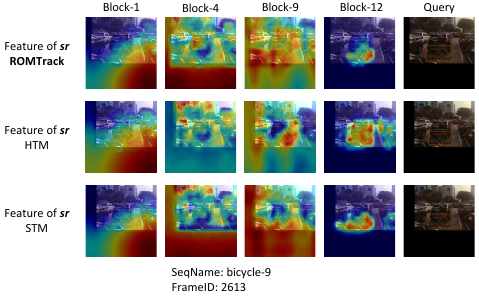}
    \caption{Visualizations of feature maps after different blocks. The \textcolor{red}{red} rectangle indicates the target object to be tracked. 
    The \textcolor{green}{green} rectangle indicates the predicted bounding box of the target object.
    Best viewed by zooming in.}
    \label{fig:feat_vis1}
\end{figure*}

\begin{figure*}
    \centering
    \includegraphics[page=6, width=.985\linewidth]{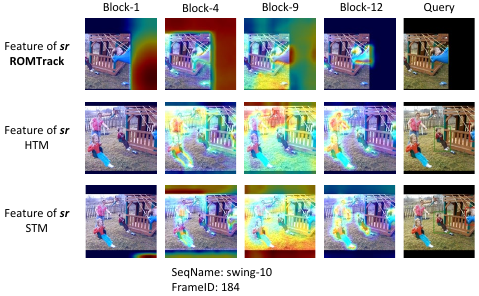}
    \includegraphics[page=8, width=.985\linewidth]{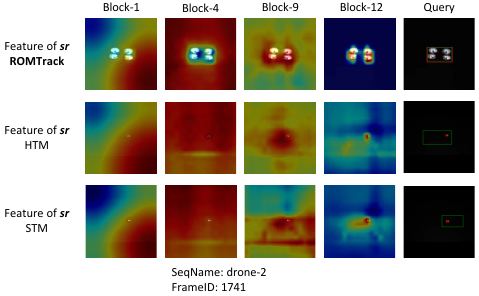}
   \caption{Visualizations of feature maps after different blocks. The \textcolor{red}{red} rectangle indicates the target object to be tracked. 
    The \textcolor{green}{green} rectangle indicates the predicted bounding box of the target object.
    Best viewed by zooming in.}
    \label{fig:feat_vis2}
\end{figure*}

\noindent \textbf{Feature Maps.} Figure~\ref{fig:feat_vis1} and Figure~\ref{fig:feat_vis2} visualize the feature maps of search regions after different blocks.
As we can see, our ROMTrack can more accurately locate the target objects, demonstrating the effectiveness of our unified attention modeling.

\end{document}